\pdfoutput=1 

\documentclass{article} % For LaTeX2e
\usepackage{iclr2020_conference,times}

\usepackage{amsmath,amsfonts,bm}

\def\eqref#1{equation~\ref{#1}}
\def\1{\bm{1}}

\def\rx{{\textnormal{x}}}

\def\rvu{{\mathbf{i}}}

\def\rvo{{\mathbf{o}}}

\def\rvr{{\mathbf{r}}}

\def\rvu{{\mathbf{u}}}
\def\rvv{{\mathbf{v}}}
\def\rvw{{\mathbf{w}}}
\def\rvx{{\mathbf{x}}}

\def\erve{{\textnormal{e}}}

\def\ervx{{\textnormal{x}}}

\def\vb{{\bm{b}}}

\def\vr{{\bm{r}}}

\def\vu{{\bm{u}}}
\def\vv{{\bm{v}}}
\def\vw{{\bm{w}}}

\def\mA{{\bm{A}}}

\def\mD{{\bm{D}}}

\def\mI{{\bm{I}}}

\def\mW{{\bm{W}}}

\DeclareMathAlphabet{\mathsfit}{\encodingdefault}{\sfdefault}{m}{sl}
\SetMathAlphabet{\mathsfit}{bold}{\encodingdefault}{\sfdefault}{bx}{n}

\usepackage{hyperref}
\usepackage{url}
\usepackage{enumerate}
\usepackage{amssymb}
\usepackage{graphicx, wrapfig}
\usepackage{booktabs}
\usepackage{subcaption}
\usepackage[normalem]{ulem}
\usepackage{multirow}

\usepackage{tikz}
\usetikzlibrary{arrows.meta, arrows,shapes, positioning, snakes, calc, shapes.geometric, decorations.pathmorphing,patterns, backgrounds}
\usetikzlibrary{tikzmark}

\newcommand{\lb}{\left(}
\newcommand{\rb}{\right)}
\newcommand{\ls}{\left[}
\newcommand{\rs}{\right]}
\newcommand{\lc}{\left\{}
\newcommand{\rc}{\right\}}
\newcommand{\setA}{\mathcal{A}}

\newcommand{\setG}{\mathcal{G}}

\newcommand{\rvphi}{\mathbf{\phi}}
\newcommand{\Real}{\mathbb{R}}

\title{Economy Statistical Recurrent Units For \\ Inferring Nonlinear Granger Causality}

\author{Saurabh Khanna \\
Department of Electrical and Computer Engineering\\
National University of Singapore\\
\texttt{elesaur@nus.edu.sg}
\And
Vincent Y. F. Tan \\
Department of Electrical and Computer Engineering\\
Department of Mathematics \\
National University of Singapore\\
\texttt{vtan@nus.edu.sg}	
}

\iclrfinalcopy % Uncomment for camera-ready version, but NOT for submission.
\begin{document}

\maketitle

\begin{abstract}
%In many applications, the networked interactions between the physical processes of interest can be qualified in terms of their pairwise Granger causal relationships. % between their respective time series measurements.  
Granger causality is a widely-used criterion for analyzing interactions in large-scale networks. As most physical interactions are inherently nonlinear, we consider the problem of inferring 
the existence of pairwise Granger causality between nonlinearly interacting stochastic processes 
from their time series measurements. Our proposed approach relies on modeling the embedded nonlinearities in the measurements using a component-wise time series prediction model based on Statistical Recurrent Units (SRUs). %and inferring the pairwise Granger causal relationships from regularized estimates of the time series fitted network weight parameters. 
%An SRU operates by extracting multi-timescale summary statistics of the past time series measurements and transforming them into highly time-sensitive predictive features. 
We make a case that the network topology of Granger causal relations is directly inferrable from a structured sparse estimate of the internal parameters of the SRU networks trained to predict the processes' time series measurements. 
We propose a variant of SRU, called \textit{economy-SRU}, which, by design  
has considerably fewer trainable parameters, and therefore less prone to overfitting. The economy-SRU computes a low-dimensional sketch of its high-dimensional hidden state in the form of random projections to generate the feedback for its recurrent processing. 
Additionally, the internal weight parameters of the economy-SRU are 
strategically regularized in a group-wise manner to facilitate the proposed network in extracting meaningful predictive features that are highly time-localized to mimic real-world causal events. 
Extensive experiments are carried out to demonstrate that the proposed economy-SRU based time series prediction model outperforms the MLP, LSTM and attention-gated CNN-based time series models considered previously for inferring Granger causality. % for model driven inference of pairwise Granger causal relationships. 
\end{abstract}

\section{Introduction}
The physical mechanisms behind the functioning of any large-scale system can be 
understood in terms of the networked interactions between the underlying system 
processes. Granger causality is one widely-accepted criterion used in building 
network models of interactions between large ensembles of stochastic processes. 
While Granger causality may not necessarily imply true causality, it has proven 
effective in qualifying pairwise interactions between stochastic processes in a 
variety of system identification problems, e.g., gene regulatory network 
mapping (\cite{Fujita07GRN}), 
and the mapping of human brain connectome (\cite{Seth15NeuroGC}). %, identifying coupling effects between climatic processes (\cite{Mosedale04ClimateGC}), etc. 
This perspective has given rise to the canonical problem of inferring pairwise Granger causal relationships between a set of stochastic processes from their time series measurements. 
At present, the vast majority of Granger causal inference methods adopt a model-based inference approach whereby the measured time series data is modeled using with a suitable parameterized data generative model whose inferred parameters ultimately reveal the true topology of pairwise Granger causal relationships. Such methods typically rely on using linear regression models for inference. %Less attention has been given to using nonlinear data generative models. 
However, as illustrated in the classical bivariate example by \cite{BaekBrock92}, 
linear model-based Granger causality tests can fail catastrophically in the presence of even mild nonlinearities in the measurements, thus making a strong case for our work which tackles the nonlinearities in the measurements %\textcolor{blue}{for Granger causal inference} 
by exploring new generative models of the time series measurements based on recurrent neural networks.

\section{Problem formulation}
\vspace{-0.1cm}
Consider a multivariate dynamical system whose evolution from an initial state is fully characterized by $n$ distinct stochastic processes which can potentially interact nonlinearly among themselves. Our goal here is to unravel the unknown nonlinear system dynamics by mapping the entire network of pairwise interactions between the system-defining stochastic processes, using Granger causality as the qualifier of the individual pairwise interactions. 

In order to detect the pairwise Granger causal relations between the stochastic processes, we assume access to their concurrent, uniformly-sampled measurements presented as an $n$-variate time series $\rvx = \lc \rvx_{t} : t \in \mathbb{N} \rc \subset \Real^{n}$. Let $\ervx_{t,i}$ denote the $i^{\text{th}}$ component of the $n$-dimensional vector measurement~$\rvx_{t}$, representing the measured value of process $i$ at time $t$.     
%Let $\rvx_{t} = \ls \ervx_{1,t}, \ervx_{2,t}, \ldots \ervx_{n,t} \rs^{T} \in \Real^{n}$ denote the cross-sectional snapshot of an $n$-dimensional time series $\rvx = \lc \rvx_{1}, \rvx_{2}, \ldots \rc$ at time instant $t$ whereby each dimension depicts uniformly sampled sequence of measurements of the $n$ distinct stochastic processes whose pairwise Granger causal relationships are being investigated.   
% Let $\rvx = \lc \ervx_{1,[t]}, \ervx_{2,[t]}, \ldots \ervx_{n,[t]} \rc$  denote the $n$-dimensional stationary time series whose $i^{\text{th}}$ component series $\ervx_{i,[t]} = \lc \rx_{i,1}, \rx_{i,2}, \ldots, \rc$ comprises uniformly sampled measurements of the $i^{\text{th}}$ process out of the $n$ distinct stochastic processes whose pairwise Granger causal relationships are being investigated.
%In order to model the nonlinear interactions between the stochastic processes, we assume that the time series measurements $\rvx_{t}$ for $t \ge 1$ are generated according to the following component-wise autoregressive generative model:  
%With inclination towards model-based inference and 
Motivated by the framework proposed in \cite{TankFox11Nips}, we assume that the measurement samples~$\rvx_{t}, 
t \in \mathbb{N}$ are generated sequentially according to the following nonlinear, component-wise autoregressive model:  
\begin{align}
\ervx_{t,i} = f_{i} \lb \rvx_{t-p:t-1, 1}, \rvx_{t-p:t-1, 2}, \ldots, \rvx_{t-p:t-1, n} \rb + \erve_{t,i}, 
\;\;\;\; i = 1, 2, \ldots n, 
\label{ts_abstract_generative_model}
\end{align}
where $\rvx_{t-p:t-1, j} \triangleq \lc \ervx_{t-1, j}, \ervx_{t-2, j}, \ldots, \ervx_{t-p, j} \rc$ represents the most recent $p$ measurements of the $j^{\text{th}}$ component of $\rvx$ in the immediate past relative to current time $t$. The scalar-valued component generative function $f_{i}$ captures all of the linear and nonlinear interactions between the $n$ stochastic processes up to time $t-1$ that decide the measured value of the $i^{\text{th}}$ stochastic process at time~$t$. The residual $\erve_{i,t}$ encapsulates the combined effect of all instantaneous and exogenous factors influencing the measurement of process $i$ at time~$t$, 
as well as any imperfections in the presumed model. Equation~\ref{ts_abstract_generative_model} may be viewed as a generalization of the linear vector autoregressive (VAR) model in the sense that the 
components of $\rvx$ can be nonlinearly dependent on one another across time.
%To ensure stationarity of $\rvx$, the generative functions $f_{i}, 1\le i \le n$ are assumed to be invariant to all past measurements beyond a certain fixed time lag $p$. Consequently, $\ervx_{i,<t}$ in \eqref{ts_abstract_generative_model} is effectively equivalent to the truncated set $\lc \ervx_{i,t-1}, \ervx_{i,t-2}, \ldots, \ervx_{i,t-p} \rc$. 
The value $p$ is loosely interpreted to be the {\em order} of the above nonlinear autoregressive model.

\vspace{-0.1cm}
\subsection{Granger causality in nonlinear dynamical systems} 
\vspace{-0.1cm}
We now proceed to interpret Granger causality in the context of the above component-wise time series model. 
Recalling the standard definition by \cite{Granger69}, a time series $\rvv$ is said to \textit{Granger cause} another time series $\rvu$ if the past of $\rvv$ contains new information above and beyond the past of~$\rvu$ that can improve the predictions of current or future values of $\rvu$. 
%The notion of Granger causality first propounded for linear VAR models by \cite{Granger69} can be extended to nonlinear autoregressive models in the following natural way. 
For $\rvx$ with its $n$ components generated according to \eqref{ts_abstract_generative_model}, the concept of Granger causality can be extended as suggested by \cite{Tank14NeuralGC} as follows.
We say that series~$j$ \textit{does not Granger cause} series $i$ if the component-wise generative function $f_{i}$ does not depend on the past measurements in series~$j$, i.e., for all $t \ge 1$ and all distinct pairs $\rvx_{t-p:t-1, j}$ and $\rvx^{\prime}_{t-p:t-1, j}$,
\begin{equation}
\label{defn_nonlin_gc}
f_{i} \lb  \rvx_{t-p:t-1, 1}, \ldots, \rvx_{t-p:t-1, j}, \ldots, \rvx_{t-p:t-1, n}\rb 
= f_{i} \lb  \rvx_{t-p:t-1, 1}, \ldots, \rvx_{t-p:t-1, j}^{\prime}, \ldots, \rvx_{t-p:t-1, n}\rb.
\end{equation}
From \eqref{ts_abstract_generative_model}, it is immediately evident that under 
the constraint in \eqref{defn_nonlin_gc},
%the current or future values of time series $i$ evolve independently irrespective of the past measurements in series $j$. In other words, 
the past of series $j$ does not assert any causal influence on series $i$, in alignment with the core principle behind Granger causality. 
Based on the above implication of \eqref{defn_nonlin_gc}, the detection of Granger noncausality between the components of $\rvx$ translates to identifying those components of $\rvx$ whose past is irrelevant to the functional description of each individual $f_{i}$ featured in \eqref{ts_abstract_generative_model}. 

Note that any reliable inference of pairwise Granger causality between the components of~$\rvx$ is feasible only if there are no unobserved confounding factors in the system which could potentially influence $\rvx$. 
In this work, we assume that the system of interest is \textit{causally sufficient} (\cite{Spirtes2016}), i.e., none of the $n$ stochastic processes (whose measurements are available) have a common Granger-causing-ancestor that is unobserved.

\vspace{-0.1cm}
\subsection{Inferring Granger causality using component-wise recurrent models}
\vspace{-0.1cm}
We undertake a model-based inference approach wherein the time series measurements are used as observations to learn an autoregressive model which is anatomically similar to the component-wise generative model described in \eqref{ts_abstract_generative_model} except for the unknown functions $f_{i}$ replaced with their respective parameterized approximations denoted by $g_{i}$. Let $\Theta_{i}, 1 \le i \le n$ denote the complete set of parameters encoding the functional description of the approximating functions~$\{g_{i}\}_{i=1}^{n}$. Then, the pairwise Granger causality between series $i$ and the rest of the components of $\rvx$ is deduced from~$\Theta_{i}$ which is estimated by fitting $g_{i}$'s output to the ordered measurements in series~$i$. 
Specifically, if the estimated $\Theta_{i}$ suggests that $g_{i}$'s output is independent 
of the past measurements in series~$j$, then we declare that series $j$ is Granger noncausal for series $i$. 
%The approximating functions~$g_{i}$ typically belong to the same function class $\setG$. 
We aim to design the approximation function~$g_{i}$ to be highly expressive and capable of well-approximating any intricate causal coupling between the components of $\rvx$ induced by the component-wise function~$f_{i}$, while simultaneously being easily identifiable from underdetermined measurements. 

By virtue of their universal approximation property (\cite{Schafer06RNNUnivApprox}), recurrent neural networks or RNNs are a particularly ideal choice for $g_{i}$ towards inferring the pairwise Granger causal relationships in $\rvx$. % in serving as substitutes for the unknown functions $f_{i}$ in the presumed generative model of $\rvx_{t}$.  
%RNNs offer a versatile yet unified framework to model and analyze the temporal dynamics of any multivariate sequential data such as text, music, videos, or in our case, a time series. 
%Gated RNNs such as long short-term memory (LSTM) and gated recurrent units (GRUs) have proved exceptionally effective in modeling complex nonlinear dependencies embedded in sequential data such as text, music, videos or a time series. 
In this work, we investigate the use of a special type of RNN called the \textit{statistical recurrent unit} (SRU) for inferring pairwise Granger causality between multiple nonlinearly interacting stochastic processes. %To this end, we propose to implement each of the $n$ approximating functions $g_{i}, 1 \le i \le n$ as distinct SRU networks. 
%serving as substitutes for the unknown functions $f_{i}$ in \eqref{ts_abstract_generative_model} 
Introduced by \cite{Oliva17SRU}, an SRU is a highly expressive recurrent neural network designed specifically for modeling multivariate time series data with complex-nonlinear dependencies spanning multiple time lags. 
%The SRU's unique multi-timescale architecture enables it to learn lag-sensitive causal features from the past time series measurements, which can be used to robustly predict the future time series values. 
Unlike the popular gated RNNs (e.g., long short-term memory (LSTM) (\cite{Hochreiter97LSTM}) and gated recurrent unit (GRU)) (\cite{Chung2014GRU}), the SRU's design is completely devoid of the highly nonlinear sigmoid gating functions and thus less affected by the vanishing/exploding gradient issue during training. Despite its simpler ungated architecture, an SRU can model both short and long-term temporal dependencies in a multivariate time series. It does so by maintaining multi-time scale summary statistics of the time series data in the past, which are preferentially sensitive to different older portions of the time series $\rvx$. By taking appropriate linear combinations of the summary statistics at different time scales, an SRU is able to construct predictive causal features which can be both highly component-specific and lag-specific at the same time. 
From the causal inference perspective, this dual-specificity of the SRU's predictive features is its most desirable feature, as one would argue that causal effects in reality also tend to be highly localized in both space and time.

\ifdefined \SKIPTEMP
In an isolated multivariate dynamical system, it is typical that the current state of a variable is causally effected by the past states of only a few other parent variables belonging to the same system. This viewpoint fosters an important assumption that the graphical topology of the pairwise Granger causality in a causally sufficient, multivariate system is sparsely connected. %For a high-dimensional VAR time series, \cite{Bahadori13} and \cite{Bolstad11GroupSparse} have shown that the assumption of sparse connectivity of the underlying Granger causal network is critical for its consistent estimation from the underdetermined time series measurements. 

For the causal inference approach using RNNs, this assumption hints towards fitting an RNN based model whose predicted output sequence is sensitive to input past measurements of only a few components of $\rvx_{t}$. This can be accomplished by strategically regularizing the weight parameters of the RNN. In \cite{Tank14NeuralGC}, .. 
\fi

\ifdefined \SKIPTEMP
% regularized function class $\setG$ have been investigated in the literature, each offering a different tradeoff between the expressiveness and ease of inference of its member functions. 
In \cite{Marrinazzo08KernelGC}, $\setG$ is taken to be the collection of functions expressible as 
a linear combination of the eigenfunctions of a certain nonlinear kernel function. 
\cite{Sun08KernelGC}, \cite{Sindhwani13KernelGC} and \cite{Lim14OKVAR} take function in $\setG$ to be expressible as sums of functions in the RKHS space induced by some matrix valued kernel. More recently, universal function approximators such as feedforward and recurrent neural networks have been considered as a proxy for $f_{i} (1 \le i \le n$) in \eqref{ts_abstract_generative_model}. 
The approximating functions $g_{i}$ are implemented as multilayer perceptron (MLP) networks or long short-term memory (LSTM) networks in \cite{Tank14NeuralGC}, and as echo state RNNs in \cite{Duggento19EchoStateForGCI}. 
\fi

%\subsection{Paper's contributions and outline}
The main contributions of this paper can be summarized as follows: 
\begin{enumerate}
	\item We propose the use of statistical recurrent units (SRUs) for detecting pairwise Granger 
	causality between the nonlinearly interacting stochastic processes. % in a dynamical system. 
	%The nonlinear interactions behind the generation of each individual process are modeled using dedicated SRUs. 
	We show that the entire network of pairwise Granger causal relationships can be inferred directly from 
	the regularized block-sparse estimate of the input-layer weight parameters of the SRUs trained to predict the time series measurements of the individual processes. 
	
	\item We propose a modified SRU architecture called \textit{economy SRU} or \textit{eSRU} in short. 
	The first of the two proposed modifications is aimed at substantially reducing the number of trainable parameters in the standard SRU model without sacrificing its expressiveness. The second modification entails regularizing the SRU's internal weight parameters to enhance the interpretability of its learned predictive features. Compared to the standard SRU, the proposed eSRU model is considerably less likely to overfit the time series measurements.
	%when trained using very few time series measurements.  
	
	\item We conduct extensive numerical experiments to demonstrate that eSRU is a compelling model for inferring pairwise Granger causality. The proposed model is found to outperform the multi-layer perceptron (MLP), LSTM and attention-gated convolutional neural network (AG-CNN) based models considered in the earlier works. %, as shown in Section~\ref{sec:sim_results}. % 
	%in \cite{Tank14NeuralGC}.      
\end{enumerate}

\ifdefined \SKIPTEMP
The rest of the paper is organized as follows. In Section~\ref{sec:proposed_method}, we propose 
a time series prediction model of $\rvx_{t}$ built using multiple SRU networks. We also motivate how 
the pairwise Granger causal relationships in $\rvx$ can be inferred by matching the output sequence of the proposed recurrent model with $\rvx_{t}$ while simultaneously enforcing group-sparsity of the model's input layer weight parameters. Next, we proceed to redesign the feedback and output feature generation paths of the original SRU network resulting in a new simpler RNN called economy SRU. Sections~\ref{sec:sim_results} and \ref{sec:experimental_results} discusses the results of our simulations conducted using synthetic datasets. and...       
Final remarks and direction for future extensions of this work are discussed in Section~\ref{sec:conclusions}.
\fi

\vspace{-0.1cm}	
\section{Proposed Granger causal inference framework} \label{sec:proposed_method}
\begin{figure*}[h]
	\begin{minipage}[c]{0.86\textwidth}	
		\resizebox{\textwidth}{!}{	
			\begin{tikzpicture}
			\begin{scope}
			%\node (dummy) at (0,-1){}; % for shifting diagram to right
			% SRU cell boundary
			\draw[dashed] (0,-0.25) rectangle ++(14.6,7);
			% RELU for input + feedback --> recurrent features
			\node[rectangle, draw, minimum height=0.75cm, minimum width=1.5cm] (RELU1) at (3, 5) {ReLU};
			% RELU for summary stats --> output features
			\node[rectangle, draw, minimum height=0.75cm, minimum width=1.5cm] (RELU2) at (10.5,5) {ReLU};		
			% Feedback RELU
			\node[rectangle, draw, minimum height=0.75cm, minimum width=1.5cm] (RELU3) at (3.5,1) {ReLU};		
			% Delay
			\node[rectangle, draw, minimum height=0.75cm, minimum width=1cm] (DLY) at (6.5,1) {$z^{-1}$};		
			% Linear output layer
			\node[rectangle, draw, minimum height=0.75cm, minimum width=1.5cm] (OUTL) at (13.5,5) {Linear};		
			% Concatenate
			\node[rectangle, draw, minimum height=0.5cm, minimum width=4cm, rotate=270] (CATL) at (8,4) {C o n c a t e n a t e};		
			\node[rectangle, draw, minimum height=0.5cm, minimum width=2cm, rotate=270] (sCATL) at (0.5,5) {Concatenate};		
			% IIR filter bank
			\node[rectangle, draw, minimum height=0.75cm, minimum width=1.5cm] (IIR1) at (6,5.5) {$H_{\alpha_{1}}(z)$};
			\node[rectangle, draw, minimum height=0.75cm, minimum width=1.5cm] (IIR2) at (6,4.5) {$H_{\alpha_{2}}(z)$};
			\node[rectangle, draw, minimum height=0.75cm, minimum width=1.5cm] (IIRm) at (6,2.5) {$H_{\alpha_{m}}(z)$};
			\node at (6, 3.5) {$\huge{\vdots}$};	
			
			% Block connections
			\draw[->] (DLY.west) to (RELU3.east);
			\draw[->] (RELU2.east) to (OUTL.west);
			\draw[->] (8.25,5) to (RELU2.west);
			\draw[->] (8.25,5) to (RELU2.west);	
			
			\draw[->] (6.75,5.5) to node[midway, above] {$\rvu_{i,t}^{\alpha_{1}}$} (7.75,5.5) ;
			\draw[->] (6.75,4.5) to node[midway, above] {$\rvu_{i,t}^{\alpha_{2}}$} (7.75,4.5);
			\draw[->] (6.75,2.5) to node[midway, above] {$\rvu_{i,t}^{\alpha_{m}}$} (7.75,2.5);
			\draw[->] ($(RELU2.south) - (1.5cm,-0.36cm)$) |- node[near end, above] {$\rvu_{i,t}$} (DLY.east);
			\coordinate[right=1.1cm of RELU1.east] (mid1);
			\coordinate[left=1.5cm of sCATL.west] (startpnt);
			\coordinate[right=1.2cm of OUTL.east] (endpnt);
			\coordinate[left=3cm of RELU3.west] (mid2);
			\draw (RELU1.east) -- node[midway, above] {$\rvphi_{i,t}$} (mid1);	
			\draw[->] (mid1) |- (IIR1.west);
			\draw[->] (mid1) |- (IIR2.west);
			\draw[->] (mid1) |- (IIRm.west);
			\draw[->] (OUTL.east) to node[near end, above] {$\hat{\ervx}_{i,t+1}$} (endpnt);	
			\draw[->] ($(startpnt) - (0,0.5cm)$) to node[near start, above] {$\rvx_{t}$} ($(sCATL.south) + (0,0.5cm)$);	
			\draw[->] (sCATL) to node[midway, above] {$\lc \rvx_{t}, \rvr_{i,t} \rc$} (RELU1.west);	
			\draw (RELU3.west) -- node[near start, above] {$\rvr_{i,t}$} (mid2) ;	
			\draw[->] (mid2) |- ($(sCATL.south) - (0,0.5cm)$);
			%Labels
			\node (output) at (12,5.2) {$\rvo_{i,t}$};
			\node (output) at (9,5.2) {$\rvu_{i,t}$};	
			\node (output) at (5.25,1.2) {$\rvu_{i,t-1}$};
			
			\node [below right=0.2cm and -1.8cm of RELU1] {$\mW_{\text{in}}^{(i)}, \mW_{\text{f}}^{(i)}, \vb_{\text{o}}^{(i)}$};		
			\node [below right=0.2cm and -1.5cm of RELU2] {$\mW_{\text{o}}^{(i)}, \vb_{\text{o}}^{(i)}$};		
			\node [below right=0.2cm and -1.5cm of RELU3] {$\mW_{\text{r}}^{(i)}, \vb_{\text{r}}^{(i)}$};	
			
			\draw[->] ($(RELU1.south) - (0.25,0.25)$) -- ($(RELU1.north) + (0.25,0.25)$);
			\draw[->] ($(RELU2.south) - (0.25,0.25)$) -- ($(RELU2.north) + (0.25,0.25)$);
			\draw[->] ($(RELU3.south) - (0.25,0.25)$) -- ($(RELU3.north) + (0.25,0.25)$);
			
			\node at (12,1.2) {\large $\displaystyle H_{\alpha}(z) := \frac{\alpha}{1 - (1-\alpha)z^{-1}}$};			
			
			\end{scope}
			\end{tikzpicture}
		}
	\end{minipage}
	\begin{minipage}[c]{0.13\textwidth}
		\caption{The $i^{\text{th}}$ SRU approximating $f_{i}$ in \eqref{ts_abstract_generative_model}. \label{fig_SRU_network}}
	\end{minipage}
	\vspace{-0.1cm}
\end{figure*}
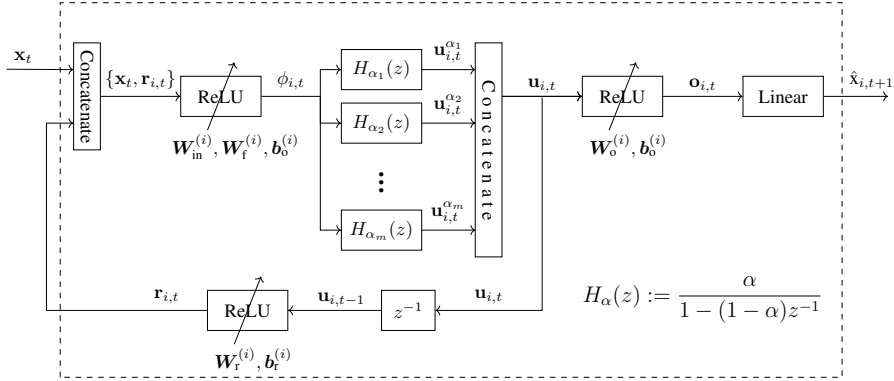

\vspace{-0.1cm}
In the proposed scheme, each of the unknown generative functions $f_{i}, 1 \le i \le n$ in the presumed component-wise model of $\rvx$ in (\ref{ts_abstract_generative_model}) is individually approximated 
by a distinct SRU network.  
The $i^{\text{th}}$ SRU network sequentially processes the time series measurements $\rvx$ and outputs a next-step prediction sequence $\hat{\ervx}_{i}^{+} = \lc \hat{\ervx}_{i,2}, \hat{\ervx}_{i,3}, \ldots, \hat{\ervx}_{i,t+1}, \ldots \rc \subset \Real$, where $\hat{\ervx}_{i,t+1}$ denotes the predicted value of component series $i$ at time~$t+1$. 
%computed recurrently using the input time series samples $\vx_{1:t} = \lc \vx_{1}, \vx_{2}, \ldots, \vx_{t} \rc$.
The prediction $\hat{\ervx}_{i,t+1}$ is computed in a recurrent fashion by combining the current input sample $\rvx_{t}$ at time $t$ with the summary statistics of past samples of $\rvx$ up to and including time~$t-1$ as illustrated in Figure~\ref{fig_SRU_network}.
%The summary statistics, denoted by $\rvu_{i,t}$, is maintained as the hidden state of the $i^{\text{th}}$ SRU network; it comprises exponentially weighted running averages of the recurrent statistics~$\rvphi_{i,t}$ evaluated for multiple different timescale values. The so called recurrent statistic $\rvphi_{t}$ is generated as the output of a single/multi layer neural network that simultaneously processes the current input $\rx_{t}$ and the feedback $\rvr_{t}$ generated from the previous hidden state/summary statistics~$\rvu_{t}$. 

The following update equations describe the sequential processing of the input time series $\rvx$ within the $i^{\text{th}}$ SRU network in order to generate a prediction of $\ervx_{i,t+1}$.  
\vspace{-0.2cm}
\begin{subequations}
	\allowdisplaybreaks		
	\begin{flushleft}
		\begin{align}
		\text{Feedback:} \; \rvr_{i,t} &= h \lb \mW^{(i)}_{\mathrm{r}} \rvu_{i,t-1} + \vb^{(i)}_{\mathrm{r}} \rb 
		\; \in \Real^{d_{\mathrm{r}}}.
		\label{eqn_sru_r_update}
		\\
		%--------------------
		\text{Recurrent statistics:} \; \rvphi_{i,t} &= h\lb \mW_{\mathrm{in}}^{(i)} \rvx_{t} + \mW_{\mathrm{f}}^{(i)} \rvr_{i,t-1} + \vb_{\mathrm{in}}^{(i)} \rb \; \in \Real^{d_{\phi}}. 
		\label{eqn_sru_phi_update}
		\\
		%--------------------
		\text{Multi-scale summary statistics:} \; \rvu_{i,t} &= \ls \lb \rvu^{\alpha_{1}}_{i,t} \rb^{T} \lb \rvu^{\alpha_{2}}_{i,t} \rb^{T} \ldots \lb \rvu^{\alpha_{m}}_{i,t} \rb^{T}  \rs^{T}  \in \Real^{m d_{\phi}},  \alpha_{j} \in \setA, \forall j.
		\label{eqn_sru_ucat}
		\\
		%--------------------
		\text{Single-scale summary statistics:} \; \rvu^{\alpha_{j}}_{i,t} &= (1-\alpha_{j}) \rvu^{\alpha_{j}}_{i,t-1} + \alpha_{j} \rvphi_{i,t}, \;\; \in \Real^{d_{\phi}},  \alpha_{j} \in [0,1].
		\label{eqn_sru_u_update}
		\\
		%--------------------
		\text{Output features:}  \; \rvo_{i,t} &= h \lb \mW_{\mathrm{o}}^{(i)} \rvu_{i,t} + \vb_{\mathrm{o}}^{(i)} \rb \; \in \Real^{d_{\mathrm{o}}}.
		\label{eqn_sru_o_update}
		\\
		%--------------------
		\text{Output prediction:} \;  \hat{\ervx}_{i,t+1} &= \lb \vw^{(i)}_{\mathrm{y}} \rb^{T} \rvo_{i,t} + b_{\mathrm{y}}^{(i)} \; \in  \Real.
		\label{eqn_sru_y_update}
		\end{align}
	\end{flushleft}
\end{subequations}
\vspace{-0.1cm}
The function $h$ in the above updates is the elementwise Rectified Linear Unit (ReLU) operator, $h(\cdot) := \max(\cdot, 0)$, which serves as the nonlinear activation in the three dedicated single layer neural networks that generate the recurrent statistics $\rvphi_{i,t}$, the feedback $\rvr_{i,t}$ and the output 
features~$\rvo_{i,t}$ in the $i^{\text{th}}$ SRU network. %The recurrent statistics $\rvphi_{i,t}$, feedback $\rvr_{i,t}$ and output features~$\rvo_{i,t}$ are $d_{\phi}, d_{\mathrm{r}}$ and $d_{\mathrm{o}}$ dimensional, real-valued vectors, respectively.
In order to generate the next-step prediction of series $i$ at time $t$, the $i^{\text{th}}$ SRU network first prepares the feedback $\rvr_{i,t}$ by nonlinearly transforming its last hidden state~$\rvu_{i,t-1}$. As stated in \eqref{eqn_sru_r_update}, a single layer ReLU network parameterized by weight matrix $\mW_{\mathrm{r}}^{(i)}$ and 
bias~$\vb_{\mathrm{r}}^{(i)}$ maps the hidden state $\rvu_{i,t-1}$ to the feedback $\rvr_{i,t}$. Another single layer ReLU network parameterized by weight matrices $\mW_{\mathrm{in}}^{(i)}, \mW_{\mathrm{f}}^{(i)}$ and bias $\vb_{\mathrm{in}}^{(i)}$ takes the input $\rvx_{t}$ and the feedback~$\rvr_{i,t}$ and tranforms them into the recurrent statistics $\rvphi_{i,t}$ as described in \eqref{eqn_sru_phi_update}. %The recurrent statistics~$\rvphi_{i,t}$ has the same connotation as the \textit{measurement innovation} in Kalman filters (\cite{Welch95KalamanFilterIntro}), in the sense that the nonlinear transformation in \eqref{eqn_sru_phi_update} can be viewed as removing all contextual information about the past up to time~$t-1$ from the current input sample~$\vx_{t}$. 
Equation \ref{eqn_sru_u_update} describes how the network's multi-timescale hidden states~$\rvu_{i,t}^{\alpha}$ for $\alpha \in \setA = \lc \alpha_{1}, \alpha_{2}, \ldots, \alpha_{m} \rc \subset \ls 0,1 \rs$ are updated in parallel by taking exponentially weighted moving averages of the recurrent statistics~$\rvphi_{i,t}$ corresponding to $m$ different scales in $\setA$. 
A third single layer ReLU network parameterized by $\mW_{\mathrm{o}}^{(i)}$ and $\vb_{\mathrm{o}}^{(i)}$ transforms the concatenated multi-timescale summary statistics $\rvu_{i,t} = \ls (\rvu_{i,t}^{\alpha_{1}})^{T} (\rvu_{i,t}^{\alpha_{2}})^{T} \ldots (\rvu_{i,m}^{\alpha_{m}})^{T}\rs^{T}$ to generate the nonlinear causal features~$\rvo_{i,t}$ which, according to \cite{Oliva17SRU}, are arguably highly sensitive to the input time series measurements at specific lags. 
Finally, the network generates the next-step prediction of series~$i$ as $\hat{\ervx}_{i,t+1}$ by linearly combining the nonlinear output features in~$\rvo_{i,t}$, as depicted in \eqref{eqn_sru_y_update}.  

%Figure~\ref{fig_SRU_network} illustrates the sequential processing of the input time series $\rvx$ inside the $i^{\text{th}}$ SRU network ongoing up to time $t$ in order to predict the next sample of series $i$ at time $t+1$. 

For values of scale $\alpha \approx 1$, the single-scale summary statistic $\rvu_{i,t}^{\alpha}$ in \eqref{eqn_sru_u_update} is more sensitive to the recent past measurements in $\rvx$. % rather than the long term past measurements. 
On the other hand, $\alpha \approx 0$ yields a 
summary statistic that is more representative of the 
older portions of the input time series. 
\cite{Oliva17SRU} elaborates on how the SRU is able 
to generate output features ($\rvo_{i,t}, 1 \le i 
\le n$) that are preferentially sensitive to the 
measurements from specific past segments of $\rvx$ 
by taking appropriate linear combinations of the 
summary statistics corresponding to different values 
of $\alpha$ in $\setA$. 

\vspace{-0.1cm}
\subsection{Inferring pairwise Granger causality using SRUs}
\vspace{-0.1cm}
Let $\Theta_{\text{SRU}}^{(i)} \triangleq \{ \mW_{\mathrm{f}}^{(i)}, \mW_{\mathrm{in}}^{(i)}, \vb_{\mathrm{in}}^{(i)}, \mW_{\mathrm{r}}^{(i)}, \vb_{\mathrm{r}}^{(i)}, \mW_{\mathrm{o}}^{(i)}, \vb_{\mathrm{o}}^{(i)}, \vw_{\mathrm{y}}^{(i)}, b_{\mathrm{y}}^{(i)} \}$ denote the complete set of parameters 
of the $i^{\text{th}}$ SRU network approximating $f_{i}$ in the presumed component-wise model of $\rvx$. From \eqref{eqn_sru_phi_update}, we observe that the weight matrix $\mW_{\text{in}}^{(i)}$ regulates 
the influence of the individual components of the input time series~$\rvx$ on the generation of the recurrent statistics $\rvphi_{i,t}$, and ultimately the next-step prediction of series $i$. 
In real-world dynamical systems, the networked interactions are typically sparse which implies that very few dimensions of the input time series $\rvx$ actually play a role in the generation of its 
individual components. Bearing this property of the networked interactions in mind, we are interested in learning the parameters~$\Theta_{\text{SRU}}^{(i)}$ such that the $i^{\text{th}}$ SRU's sequential output closely matches with series $i$'s measurements, while simultaneously seeking a \textit{column-sparse} estimate of the weight matrix $\mW_{\mathrm{in}}^{(i)}$. %, i.e., majority of the columns in the estimated $\mW_{\mathrm{in}}^{(i)}$ contain only zero elements. 

We propose to learn the parameters $\Theta_{\text{SRU}}^{(i)}$ of the $i^{\text{th}}$ SRU 
network by minimizing the penalized mean squared prediction error loss as 
shown below. 
\begin{equation}
\hat{\Theta}^{(i)}_{\text{SRU}} := \underset{\Theta^{(i)}_{\text{SRU}}}{\arg \min} \; \frac{1}{T-1} \sum_{t = 1}^{T-1} \lb \hat{\ervx}_{i,t} - \ervx_{i,t+1} \rb^{2} + \lambda_{1} \sum_{j = 1}^{n} \| \mW^{(i)}_{\mathrm{in}}(:,j) \|_{2}.
\label{eqn_SRU_basic_opt}
\end{equation}
In the above, the network output $\hat{\ervx}_{i,t}$ depends nonlinearly on $\mW_{\mathrm{in}}^{(i)}$ according to the composite relation described by the updates 
(\ref{eqn_sru_r_update})-(\ref{eqn_sru_y_update}) and $\mW^{(i)}_{\mathrm{in}}(:,j)$ 
denotes the $j^{\text{th}}$ column in the weight matrix 
$\mW^{(i)}_{\mathrm{in}}$. The~$\ell_{1}$-group norm penalty in the objective is known 
to promote column sparsity in the estimated $\mW_{\mathrm{in}}^{(i)}$ 
(\cite{Simon13GrpLasso}). 
From \eqref{eqn_sru_phi_update}, a straightforward implication of the column vector $\mW^{(i)}_{\text{in}}(:,j)$ 
being estimated as the all-zeros vector is that the past measurements in series~$j$ 
do not influence the predicted future value of series~$i$. In this case, we declare that series $j$ does not Granger-cause series $i$. Moreover, the index set supporting the non-zero columns in the 
estimated weight matrix $\hat{\mW}_{\text{in}}^{(i)}$ enumerates the components of~$\rvx$ which are likely to Granger-cause series $i$. Likewise, the entire network of pairwise Granger causal relationships in $\rvx$ can be deduced from the non-zero column support of the estimated weight matrices $\mW_{\text{in}}^{(i)}, 1 \le i \le n$ in the~$n$ SRU networks trained to predict the components of $\rvx$.

The component-wise SRU optimization problem in \eqref{eqn_SRU_basic_opt} is nonconvex and potentially has multiple local minima. To solve for $\hat{\Theta}_{\text{SRU}}^{(i)}$, we use first-order gradient-based methods such as stochastic gradient descent which have been found to be consistently successful in finding good solutions of nonconvex deep neural network optimization problems (\cite{AllenZhu19DeepOpt}). Since our approach of detecting Granger noncausality hinges upon correctly identifying the all-zero columns 
of~$\mW_{\mathrm{in}}^{(i)}$, it is important that the first-order gradient based parameter updates used for minimizing the penalized SRU loss ensure that majority of the coefficients in $\mW_{\mathrm{in}}^{(i)}$ iterates become exactly zero after a certain number of iterations. Seeking exact column sparsity in the converged solution of~$\mW_{\mathrm{in}}^{(i)}$, we follow the same approach as \cite{Tank14NeuralGC} and resort to a first-order proximal gradient descent algorithm to find a regularized solution of the SRU optimization. 
%The proximal gradient operator associated with the $\ell_{1}$-group norm penalty term in \eqref{eqn_SRU_basic_opt} is a columwise soft-thresholding operator which is applied to the $\mW_{\mathrm{in}}^{(i)}$ iterate after every standard gradient descent update, until convergence. 
The gradients needed for executing the gradient descent updates of the SRU network parameters are computed efficiently using the \textit{backpropagation through time} (BPTT) procedure (\cite{Jaeger02TutBPTT}). 
%such as stochastic gradient descent, Adam, AdaGrad and RMSProp 

\vspace{-0.1cm}
\section{Economy SRU: A remedy for overfitting}
\vspace{-0.1cm}
By computing the summary statistics of past measurements at sufficiently granular time scales, an SRU can learn predictive causal features %, namely $\rvo_{i,t}, 1 \le i \le n$ 
which are highly localized in time. While a higher granularity of $\alpha$ in $\setA$ translates to a more general SRU model that fits better to the 
time series measurements, it also entails substantial increase in the number of trainable parameters.
% potentially increasing the chances of overparameterized model. 
Since measurement scarcity is typical in causal inference problems, the proposed component-wise SRU based time series prediction model is usually overparameterized and thus susceptible to overfitting.
%The largest contributors of trainable parameters in the $i^{\text{th}}$ SRU network are the weight matrices $\mW_{r}^{(i)}$ and $\mW_{o}^{(i)}$ associated with the ReLU networks generating the feedback and the output features, respectively. The number of weight coefficients in $\mW_{r}^{(i)}$ and $\mW_{o}^{(i)}$ scale linearly with $m$, the number of distinct timescales in $\setA$ used to compute the multi-timescale summary statistics. 
The typical high dimensionality of the recurrent statistic~$\rvphi_{t}$ accentuates this issue. %An overparameterized SRU model is susceptible to overfitting, especially so, when our goal is to infer Granger causal relationships from as few time series measurements as possible. 
%In the following section, we discuss the measures to address the overparameterization and overfitting related issues faced by the SRU based component-wise time series model. 

%\subsection{Economy Statistical Recurrent Units ($e$SRU)}
To alleviate the overfitting concerns, we propose two modifications to the standard SRU (\cite{Oliva17SRU}) aimed primarily at reducing its 
likelihood of overfitting the time series measurements. %While the first modification directly reduces the number of trainable weight coefficients in $\mW_{r}^{(i)} (1 \le i \le n)$, the second modification relies on meaningful regularization of the weight matrices $\mW_{o}^{(i)}$.
The modifications are relevant regardless of the current Granger causal inference context, and henceforth we refer to the modified SRU as \textit{Economy-SRU} (\textit{eSRU}).
%Compared to the standard SRU, the proposed eSRU model contains fewer trainable parameters and therefore lesser prone to overfitting. 

\vspace{-0.1cm}
\subsection{Modification-I: Generating feedback from a 
low-dimensional sketch of summary statistics} 
\label{sec:sru_mod1}
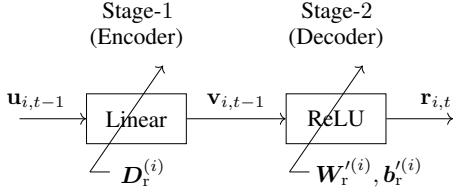
\begin{wrapfigure}{l}{6.4cm}
	\vspace{-0.1cm}		
	\begin{center}
		\resizebox{0.45\textwidth}{!}{
			\begin{tikzpicture}
			\node[rectangle, draw, minimum height=0.75cm, minimum width=1.5cm] (LIN) at (5, 2) {Linear};
			\node[rectangle, draw, minimum height=0.75cm, minimum width=1.5cm] (RELU) at (8, 2) {ReLU};		
			\draw[->] (LIN.east) to node[midway, above] {$\rvv_{i,t-1}$} (RELU.west) ;
			\coordinate [right=1cm of RELU.east] (endpnt);
			\coordinate [left=1cm of LIN.west] (startpnt);
			\draw[->] (startpnt) to node[near start, above] {$\rvu_{i,t-1} $} (LIN.west) ;
			\draw[->] (RELU.east) to node[near end, above] {$\rvr_{i,t} $} (endpnt) ;
			\node at (5.1, 1.2) {$\mD_{\mathrm{r}}^{(i)}$};
			\node at (8.5, 1.2) {$\mW_{\mathrm{r}}^{\prime (i)}, \vb_{\mathrm{r}}^{\prime (i)}$};
			\begin{scope}[on background layer]    % select the background layer
			\draw[->] (4.3, 1.2) to (5.4, 2.8);
			\draw[->] (7.3, 1.2) to (8.4, 2.8);
			\end{scope}
			\draw (4.3, 1.2) -- (4.6, 1.2);
			\draw (7.3, 1.2) -- (7.6, 1.2);
			\node at (5, 3.4) {\vtop{\hbox{\strut $\;$ Stage-$1$} \hbox{\strut (Encoder)}}};
			\node at (8, 3.4) {\vtop{\hbox{\strut $\;$ Stage-$2$} \hbox{\strut (Decoder)}}};
			\end{tikzpicture}
		}
	\end{center}	
	\caption{Proposed two-stage feedback in economy-SRU. %The trainable parameters are highlighted in red.
	}
	\label{fig_proposed_feedback_net}
	\vspace{-0.1cm}
\end{wrapfigure}

\vspace{-0.1cm}
\label{sec:two_stage_feedback}
We propose to reduce the number of trainable parameters in the $i^{\text{th}}$ SRU network by substituting the feedback ReLU network parametrized by $\mW_{\mathrm{r}}^{(i)}$ 
and~$\vb_{\mathrm{r}}^{(i)}$ with the two stage network shown in Fig.~\ref{fig_proposed_feedback_net}. 
The first stage implements the linear matrix-vector multiplication operation $\mD_{\mathrm{r}}^{(i)} \rvu_{i,t}$ to generate the output $\vv_{i,t} \in \Real^{d_{\mathrm{r}}^{\prime}}$, where $\mD_{\mathrm{r}}^{(i)} \in \Real^{d_{\mathrm{r}}^{\prime} \times m d_{\phi}}$ is a fixed, full row-rank matrix with $d_{\mathrm{r}}^{\prime} \ll m d_{\phi}$. 
The $d^{\prime}_{\mathrm{r}}$-dimensional output of the first stage can be viewed as a low-dimensional, stable embedding of the multi-timescale summary statistics $\rvu_{i,t}$. The entries of the constant matrix $\mD_{\mathrm{r}}^{(i)}$ are drawn independently from a zero mean Gaussian distribution with variance 
$\frac{1}{d_{\mathrm{r}}^{\prime}}$. The stage-$1$ processing is based on the premise that for most real-world systems and
the associated time series measurements, their high-dimensional summary statistics learned by the SRU network 
as $\rvu_{i,t}$ tend to be highly structured, and thus $\rvu_{i,t}$ has significantly fewer degrees of freedom relative to its ambient dimension. %and admit sparse representations under suitable basis transformations.
Thus, by projecting the $md_{\phi}$-dimensional 
$\rvu_{i,t}$ onto the $d_{\mathrm{r}}^{\prime} (\ll 
m d_{\phi})$ rows of 
$\mD_{\mathrm{r}}^{(i)}$, we obtain its 
low-dimensional embedding $\vv_{i,t}$ which 
nonetheless retains most of the 
contextual information conveyed by the 
uncompressed~$\rvu_{i,t}$\footnote{Gaussian
 random matrices of appropriate 
dimensions are approximately isometries with overwhelming 
probability (\cite{JLLemma84}). However, instead of 
using $n$ independent instantiations of a Gaussian 
random matrix 
for initializing $\mD_{\mathrm{r}}^{(i)}$, $1 \le i 
\le n$, we recommend initializing them with the same 
random matrix, as the latter strategy reduces the 
probability that any one of them is spurious encoder 
by $n$-fold.} 
\cite{JLLemma84, 
Dirksen2014DimensionalityRW}. 
%The intuition behind stage-$1$ processing is that $\mD_{r}^{(i)}$ sketches the high-dimensional summary statistics vector $\rvu_{i,t}$ to its much lower-dimensional embedding $\rvv_{i,t}$ which arguably retains most of the contextual information conveyed by $\rvu_{i,t}$.
The second stage of the proposed feedback network is a single/multi-layer ReLU network which maps the sketched summary statistics~$\vv_{i,t}$ to the feedback vector $\rvr_{i,t}$. The second stage ReLU network is parameterized by weight matrix $\mW_{\mathrm{r}}^{\prime, (i)} \in \Real^{d_{\mathrm{r}} \times d_{\mathrm{r}}^{\prime}}$ and bias $\vb_{\mathrm{r}}^{\prime, (i)} \in \Real^{d_{\mathrm{r}}}$. 
Compared to the standard SRU's feedback whose generation is controlled by $m d_{\phi} d_{\mathrm{r}} + d_{\mathrm{r}}$ trainable parameters, the proposed feedback network has only $d_{\mathrm{r}}^{\prime} d_{\mathrm{r}} + d_{\mathrm{r}}$ trainable parameters, which is substantially fewer when $d_{\mathrm{r}}^{\prime } \ll m d_{\phi}$. Consequently, the modified SRU is less susceptible to overfitting.

%The $i^{\text{th}}$ SRU network approximating $f_{i}$ maintains a multi-timescale summary statistic of the past measurements of $\rvx_{t}$ as the network's hidden state $\rvu_{i,t}$. The dimension of this hidden state grows linearly with the number of timescales $\alpha$ in $\setA$. Consequently, the number of trainable parameters in the single layer ReLU network mapping the summary statistics to the final predictive features in $\rvo_{i,t}$ also grows linearly with the number of timescales in $\setA$. At the same time, maintaining summary statistics at more granular timescales allows the SRU to learn causal predictive features which can be highly localized in time. This is desirable, as often in reality, interesting patterns in a time series can be causally attributed to one or more highly time-localized events in the past. % of the same or some other time series.

%\subsubsection{Modification-II: Generating time-localized predictive features via grouped-sparse mixing of multi-scale summary statistics}
\vspace{-0.1cm}
\subsection{Modification-II: Grouped-sparse mixing of 
multi-scale summary statistics for learning 
time-localized predictive features} 
\label{sec:sru_mod2}
\vspace{-0.1cm}
In the standard SRU proposed by \cite{Oliva17SRU}, there are no restrictions on the weight 
matrix~$\mW_{\mathrm{o}}^{(i)}$ parameterizing the ReLU network that maps the summary statistics $\rvu_{i,t}$ to the final predictive features in $\rvo_{i,t}$. Noting that the number of parameters in the $m d_{\mathrm{\phi}} \times d_{\mathrm{o}}$ sized weight matrix~$\mW_{\mathrm{o}}^{(i)}$ usually dominates the overall number of trainable parameters in the SRU, any meaningful effort towards addressing the model overfitting concerns must consider regularizing the weights in~$\mW_{\mathrm{o}}^{(i)}$. 
In this spirit, we propose the following penalized optimization problem to estimate the  
parameters $\Theta_{\text{eSRU}}^{(i)} = ( \Theta_{\text{SRU}}^{(i)} \backslash \{  \mW_{\mathrm{r}}^{(i)} \}  ) \cup \{  \mW_{\mathrm{r}}^{\prime (i)} \}$
%\lc \mW_{\mathrm{in}}^{(i)}, \mW_{\mathrm{f}}^{(i)}, \vb_{\mathrm{in}}^{(i)}, \mW_{\mathrm{r}}^{\prime (i)}, \vb_{\mathrm{r}}^{(i)}, \mW_{\mathrm{o}}^{(i)}, \vb_{\mathrm{o}}^{(i)}, \vw_{\mathrm{y}}^{(i)}, b_{\mathrm{y}}^{(i)} \rc$ 
of the eSRU model equipped with the two-stage feedback proposed in Section~\ref{sec:two_stage_feedback}:
%\begin{align}
%\hat{\Theta}_{\text{eSRU}}^{(i)} = \underset{\Gamma^{(i)}}{\arg \min} \; \frac{1}{T-1} 
%&\sum_{t = 1}^{T-1} \lb y_{i,t} - x_{i,t+1} \rb^{2} + \lambda_{1} \sum_{j = 1}^{n} \| \mW^{(i)}_{\mathrm{in}}(:,j) \|_{2}  
%\nonumber \\
%-----------------
%& \hspace{2cm}
%+ \lambda_{2} \sum_{j = 1}^{d_{\mathrm{o}}} \sum_{k = 1}^{d_{\phi}} 
%\sqrt{ \sum_{l = 0}^{m-1} \lb \mW_{\mathrm{o}}^{(i)} (j, (l-1)d_{\phi} + k) \rb^{2} }, 
%\label{eqn_SRU_opt_mod2}
%\end{align} 
\begin{equation}
\hat{\Theta}_{\text{eSRU}}^{(i)} = \underset{\Theta^{(i)}_{\text{eSRU}}}{\arg \min}  \frac{1}{T\!-\!1} 
\sum_{t = 1}^{T-1} \lb \hat{\ervx}_{i,t} - \ervx_{i,t+1} \rb^{2} + \lambda_{1} \sum_{j = 1}^{n} \| \mW^{(i)}_{\mathrm{in}}(:,j) \|_{2}  
+ \lambda_{2} \sum_{j = 1}^{d_{\mathrm{o}}} \sum_{k = 1}^{d_{\phi}} 
\| \mW_{\mathrm{o}}^{(i)} (j, \setG_{j,k}) \|_{2}.
\label{eqn_SRU_opt_mod2}
\end{equation} 
Here $\lambda_{1}$ and $\lambda_{2}$ are positive constants that bias the group sparse penalizations against the eSRU's fit to the measurements in the $i^{\text{th}}$ component series. 
The term $\mW_{\mathrm{o}}^{i}(j, \setG_{j,k})$ ($1 \le j \le d_{\mathrm{o}}, 1 \le k \le d_{\phi}$) denotes the subvector of the $j^{\text{th}}$ row in $\mW_{\mathrm{o}}^{(i)}$ obtained by extracting the weight coefficients indexed by set $\setG_{j,k}$.
%comprises weight coefficients in~$\mW_{\mathrm{o}}^{(i)}$ are partitioned into $d_{\phi} d_{\mathrm{o}}$ different groups $\setG_{j,k}, 1 \le j \le d_{\mathrm{o}}, 1 \le k \le d_{\phi}$. %such that each group comprises $m$ weights multiplying the $m$  
As shown via an example in Fig. \ref{fig_groups_in_Wo}, the index set $\setG_{j,k}$ enumerates the $m$ weight coefficients in the row vector $\mW_{\mathrm{o}}^{(i)}(j,:)$ which are multiplied to the exponentially weighted running averages of $k^{\text{th}}$ recurrent statistic $\rvphi_{i,t}(k)$ corresponding to the $m$ timescales in~$\setA$ prior to being transformed by the neural unit generating the $j^{\text{th}}$ predictive feature in $\rvo_{i,t}$.
Compared to \eqref{eqn_SRU_basic_opt}, the second penalty term in \eqref{eqn_SRU_opt_mod2} promotes a group-sparse solution for~$\mW_{\mathrm{o}}^{(i)}$ to the effect that each predictive feature in $\rvo_{i,t}$ depends on only a few components of the recurrent statistic $\rvphi_{i,t}$ via their linearly mixed multi-scale exponentially weighted averages. 
%\textcolor{blue}{Furthermore, this special grouping of the weights in $\mW_{\mathrm{o}}^{(i)}$ guarantees that prior to the nonlinear activation in the neural units that generate the output features in $\rvo_{i,t}$, the linear mixing of the multi-scale running averages of each individual recurrent feature in $\rvphi_{i,t}$ occurs separately. } 
We opine that the learned linear mixtures, 
represented by the intermediate 
products~$\mW_{\mathrm{o}}^{(i)}(j,\setG_{j,k}) 
\rvu_{i,t}(\setG_{k,j})$, are highly sensitive to 
certain past segments of the input time series 
$\rvx$. Consequently, the output features in 
$\rvo_{i,t}$ are both time-localized and 
component-specific, a common trait of real-world 
causal effects.  
%\cite{Oliva17SRU}, taking appropriate weighted sums 
%of the multi-timescale summary statistics, one can 
%generate time-localized features which are 
%sensitive 
%to only specific past segments of the input time 
%series. 

The above group-sparse regularization of the weight coefficients in 
$\mW_{\mathrm{o}}^{(i)}$, combined with the column-sparsity of 
$\mW^{(i)}_{\text{in}}$, is pivotal to enforcing that the occurrence of any 
future pattern in a time series can be attributed to the past 
occurrences of a few highly time-localized patterns in the ancestral time 
series.
The results of our numerical experiments further confirm that by choosing $\lambda_{1}$ and $\lambda_{2}$ appropriately, the proposed group-wise sparsity inducing regularization of $\mW_{\mathrm{o}}^{(i)}$ ameliorates overfitting, and the optimization in \eqref{eqn_SRU_opt_mod2} yields an estimate of $\mW_{\text{in}}^{(i)}$ whose column support closely reflects the true pairwise Granger causal relationships between the components of $\rvx$.
\begin{center}
	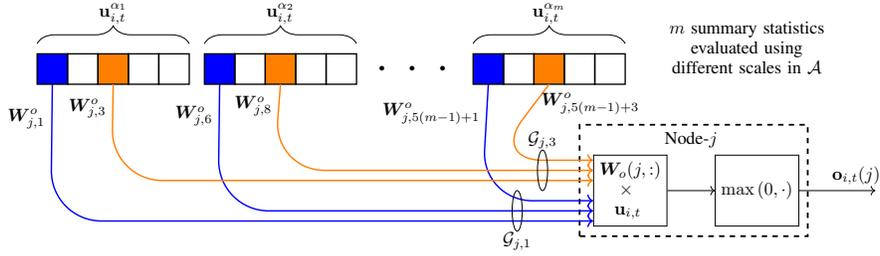
\begin{figure*}
		\centering
		\resizebox{0.85\textwidth}{!}{
			\begin{tikzpicture}
			% \origin 
			\node at (0,0) {};
			%multi-scale summary stats
			\draw[line width=0.75, fill=blue] (0.5,3) rectangle ++(0.6,0.6);
			\draw[line width=0.75] (0.5+1*0.6,3) rectangle ++(0.6,0.6);
			\draw[line width=0.75, fill=orange] (0.5+2*0.6,3) rectangle ++(0.6,0.6);
			\draw[line width=0.75] (0.5+3*0.6,3) rectangle ++(0.6,0.6);
			\draw[line width=0.75] (0.5+4*0.6,3) rectangle ++(0.6,0.6);
			\draw[line width=0.75, fill=blue] (3.8,3) rectangle ++(0.6,0.6);
			\draw[line width=0.75] (3.8+1*0.6,3) rectangle ++(0.6,0.6);
			\draw[line width=0.75, fill=orange] (3.8+2*0.6,3) rectangle ++(0.6,0.6);
			\draw[line width=0.75] (3.8+3*0.6,3) rectangle ++(0.6,0.6);
			\draw[line width=0.75] (3.8+4*0.6,3) rectangle ++(0.6,0.6);
			\node at (7.3, 3.3) {\Huge{$\cdot$}};
			\node at (7.9, 3.3) {\Huge{$\cdot$}};
			\node at (8.5, 3.3) {\Huge{$\cdot$}};
			\draw[line width=0.75, fill=blue] (9.1,3) rectangle ++(0.6,0.6);
			\draw[line width=0.75] (9.1+1*0.6,3) rectangle ++(0.6,0.6);
			\draw[line width=0.75, fill=orange] (9.1+2*0.6,3) rectangle ++(0.6,0.6);
			\draw[line width=0.75] (9.1+3*0.6,3) rectangle ++(0.6,0.6);
			\draw[line width=0.75] (9.1+4*0.6,3) rectangle ++(0.6,0.6);
			
			\node at (14.5, 3.8) [text width=3.5cm]{\begin{center}$m$ summary statistics evaluated using different scales in $\setA$ \end{center}};
			
			\draw [decorate,decoration={brace,amplitude=10pt},xshift=0pt,yshift=0pt] (0.5,3.8) -- (0.5+5*0.6,3.8) node [black,midway,yshift=0.6cm] {$\rvu_{i,t}^{\alpha_{1}}$};
			\draw [decorate,decoration={brace,amplitude=10pt},xshift=0pt,yshift=0pt] (3.8,3.8) -- (3.8+5*0.6,3.8) node [black,midway,yshift=0.6cm] {$\rvu_{i,t}^{\alpha_{2}}$};
			\draw [decorate,decoration={brace,amplitude=10pt},xshift=0pt,yshift=0pt] (9.1,3.8) -- (9.1+5*0.6,3.8) node [black,midway,yshift=0.6cm] {$\rvu_{i,t}^{\alpha_{m}}$};
			
			\draw[dashed, line width=1, black] (7+4.2, 0) rectangle ++(4.5,2.2);
			\node[rectangle, draw, minimum width=1.4cm,minimum height=1.4cm] (relu) at (7+7.7,0.9) {$\max \lb 0, \cdot \rb$};
			\node[rectangle, draw, minimum width=1.4cm,minimum height=1.4cm] (wsum) at (7+5.2,0.9) 
			{	\vtop{\hbox{\strut $\mW_{o}(j,:)$}
					\hbox{\strut $\;\;\;\;\times$}
					\hbox{\strut $\;\;\;\rvu_{i,t}$}}};
			\node at (7+6.4, 1.9) {Node-$j$};
			
			\coordinate [right=1.5cm of relu.east] (endpnt);
			\draw[->] (wsum.east) to (relu.west) ;
			\draw[->] (relu.east) to node[near end, above] {$\rvo_{i,t}(j) $} (endpnt);
			\coordinate [left=7cm of wsum.west] (grp1);
			
			\draw[->, thick, blue, rounded corners=1cm] (0.8,3) to node[near start, left] {\textcolor{black}{$\mW^{o}_{j,1}$}} ($(grp1)-(3.7,0.6)$)--($(wsum.west) - (0,0.6)$);
			\draw[->, thick, blue, rounded corners=1cm] (3.8+0.3,3) to node[near start, left] {\textcolor{black}{$\mW^{o}_{j,6}$}} ($(grp1)-(0.4,0.4)$)--($(wsum.west) - (0,0.4)$);
			\draw[->, thick, blue, rounded corners=1cm] (9.1+0.3,3) to node[near start, left] {\textcolor{black}{$\mW^{o}_{j,5(m-1) + 1}$}} ($(grp1)-(-4.8,0.2)$)--($(wsum.west) - (0,0.2)$);
			
			\draw[->, thick, orange, rounded corners=1cm] (2,3) to node[near start, left] {\textcolor{black}{$\mW^{o}_{j,3}$}} ($(grp1)+(-2.5,0.2)$)--($(wsum.west) + (0,0.2)$);
			\draw[->, thick, orange, rounded corners=1cm] (3.8+1.5,3) to node[near start, left] {\textcolor{black}{$\mW^{o}_{j,8}$}} ($(grp1)+(0.7,0.4)$)--($(wsum.west) + (0,0.4)$);
			\draw[->, thick, orange, rounded corners=1cm] (9.1+1.5,3) to node[near start, right] {\textcolor{black}{$\mW^{o}_{j,5(m-1)+3}$}} ($(grp1)+(5,0.6)$)--($(wsum.west) + (0,0.6)$);
			
			\draw ($(wsum.west) - (1.5,0.4)$) circle [x radius=0.1, y radius=0.4];
			\draw ($(wsum.west) + (-1,0.4)$) circle [x radius=0.1, y radius=0.4];
			\node at ($(wsum.west) - (1.5,1)$) {$\setG_{j,1}$};
			\node at ($(wsum.west) + (-1,1)$) {$\setG_{j,3}$};
			
			\end{tikzpicture}
		}
		\caption{An illustration of the proposed group-wise mixing of the multi-timescale summary statistics $\rvu_{i,t}$ in the $i^{\text{th}}$ SRU (with $d_{\phi}$ = $5$) towards generating the $j^{\text{th}}$ predictive feature in $\rvo_{i,t}$. The weights corresponding to the same colored connections belong to the same group.}  
		\label{fig_groups_in_Wo}
		\vspace{-0.3cm}
	\end{figure*}
\end{center}

\vspace{-0.1cm}
\section{Experiments}  \label{sec:sim_results}
\vspace{-0.1cm}
We evaluate the performance of the proposed SRU- and eSRU-based component-wise time series models in inferring pairwise Granger causal relationships in a multivariate time series. The proposed models are compared to the existing MLP- and LSTM-based models in \cite{Tank14NeuralGC} and the attention-gated CNN-based model (referred hereafter as Temporal Causal Discovery Framework (TCDF)) in \cite{Nauta19TCDF}. To ensure parity between the competing models, the maximum size of all the input/hidden/output layers in the different NN/RNN time series models is fixed to $10$, unless specified otherwise. The complete list of tuned hyperparameters of the considered models used for different datasets is provided in Appendix~\ref{app:tuned_hyperparams}. The performance of each method is qualified in terms of its AUROC (Area Under the Receiver Operating Characteristic curve). Here, the ROC curve illustrates the trade off between the true-positive rate (TPR) and the false-positive rate (FPR) achieved by the methods towards the detection of $n^2$ pairwise Granger causal relationships between the $n$ measured processes in the experiment. The ROC curves of SRU and eSRU models are obtained by sweeping through different values of the regularization parameter $\lambda_{1}$ in \eqref{eqn_SRU_basic_opt} and \eqref{eqn_SRU_opt_mod2}, respectively. Likewise, the ROCs of component-wise MLP and LSTM models are obtained by varying $\lambda_{1}$'s counterpart in \cite{Tank14NeuralGC}. For TCDF, the ROC curve is obtained by varying the 
threshold that is applied to attention scores of the trained AG-CNN model in \cite{Nauta19TCDF}.       

%The results of our experiments are discussed below.

\vspace{-0.1cm}
\subsection{Lorenz-96 simulations}
\vspace{-0.1cm}
In the first set of experiments, the time series measurements $\rvx$ intended 
for Granger causal inference are generated according to the Lorenz-96 model which has been extensively used in climate science for modeling and prediction purposes (\cite{Schneider17Lorenz96}). In the Lorenz-$96$ model of an $n$-variable system, the individual state trajectories of the $n$ variables are governed by the following set of odinary differential equations:
\begin{equation}
\label{eqn_lorenz_model}
\frac{\partial \ervx_{t,i}}{\partial t} = - \ervx_{t,i-1} \lb \ervx_{t,i-2} - \ervx_{t,i+1} \rb - \ervx_{t,i} + F, \;\;\;\; 1 \le i \le n.
\end{equation}  
where the first and the second terms on the RHS represent the \textit{advection} and the \textit{diffusion} in the system, respectively, and the third term $F$ is the magnitude of the external forcing. The system dynamics becomes increasingly chaotic for higher values of $F$ (\cite{Karimi10Lorenz}). We evaluate and compare the 
accuracy of the proposed methods in inferring pairwise Granger causal relationships between $n=10$ variables 
with Lorenz-96 dynamics. 
We consider two settings: $F=10$ and $F=40$ in order to simulate two different strengths of nonlinearity in the causal interactions between the variables. Here, the ground truth is straightforward i.e., 
for any $1 \le i \le n$, the $i^{\text{th}}$ component of time series $\rvx$ is Granger caused by its 
components with time indices from $i-2$ to $i+1$.
% $(i\!-\!2)^{\text{th}}, (i\!-\!1)^{\text{th}}, i^{\text{th}}$ and~$(i\!+\!1)^{\text{th}}$ components. 
%The  reported AUROC values are averaged over five independently generated Lorenz-96 datasets. 

%We consider a single layer design all throughout in the competing models with the number of neurons per layer fixed to $10$. 
In the case of weak nonlinear interactions ($F$ = $10$), from Table~\ref{tab:lorenz_AUROC_F10}, we observe that eSRU achieves the highest AUROC among all competing models. The gap in performance is more pronounced when fewer time series measurements ($T$ = $250$) are available.
In case of strong nonlinear interactions ($F$ = $40$), we observe that both SRU and eSRU are the only models that are able to perfectly recover the true Granger causal network (Table~\ref{tab:lorenz_AUROC_F40}).  
Surprisingly, the SRU and eSRU models perform poorer when $F$ is small. This could be attributed to the proposed models not sufficiently regularized when fitted to weakly-interacting time series measurements that are less nonlinear.

\begin{table}[t]
\begin{small}	
	\caption{Averaged AUROC for $5$ independent Lorenz-96 datasets}
	\begin{subtable}{.45\textwidth}
		\caption{$F$ = $10$}
		\label{tab:lorenz_AUROC_F10}
		\begin{tabular}{lcc}
			\multicolumn{1}{c}{\bf MODEL}  &\multicolumn{2}{c}{\bf AVERAGE AUROC}
			\\ \hline \vspace{-0.2cm} \\
			& \bf $T = 250$ & \bf $T = 500$ 
			\\ \hline \\
			MLP         	& $0.93 \pm 0.02$			& $0.96 \pm 0.03$			 \\
			LSTM 			& $0.90 \pm 0.02$			& $0.95 \pm 0.05$			 \\
			TCDF			& $0.70 \pm 0.01$           & $0.72 \pm 0.04$            \\    
			SRU				& $0.84	\pm 0.03$			& $0.90 \pm 0.02$			 \\
			eSRU  			& \bm{$0.95 \pm 0.02$}		& \bm{$0.98 \pm 0.01$}		 
			\\
		\end{tabular}
	\end{subtable}
	\hspace{1cm}
	\begin{subtable}{.45\textwidth}
		\caption{$F$ = $40$}
		\label{tab:lorenz_AUROC_F40}
		\begin{tabular}{lcc}
			\multicolumn{1}{c}{\bf MODEL}  &\multicolumn{2}{c}{\bf AVERAGE AUROC}
			\\ \hline \vspace{-0.2cm} \\
			& \bf $T = 250$ & \bf $T = 500$ 
			\\ \hline \\
			MLP 	& $0.85 \pm 0.08$				& $0.94 \pm 0.03$				 \\
			LSTM 	& $0.78 \pm 0.09$				& $0.90 \pm 0.05$				 \\
			TCDF    & $0.62 \pm 0.01$             	& $0.68 \pm 0.04$   			 \\
			SRU 	& \bm{$1.0 \pm 0.0$}			& \bm{$1.0 \pm 0.0$}			      \\
			eSRU 	& $0.99 \pm 0.0$				& \bm{$1.0 \pm 0.0$}		 \\
		\end{tabular}
	\end{subtable}
\end{small}
\vspace{-0.1cm}
\end{table}

\vspace{-0.1cm}
\subsection{VAR simulations}
\vspace{-0.1cm}
In the second set of simulations, we consider the time series measurements $\rvx$ to be generated according to a $3^{\text{rd}}$ order linear VAR model:
\begin{equation} \label{eqn_linear_var_model}
\rvx_{t} = \mA^{(1)} \rvx_{t-1} + \mA^{(2)} \rvx_{t-2} + \mA^{(3)} \rvx_{t-3} + \rvw_{t},  \;\;\;\; t \ge 1, 
\end{equation}
where the matrices $\mA^{(i)}, i = 1,2,3$ contain the regression coefficients which model the linear interactions between its $n = 10$ components. The noise term $\rvw_{t}$ is Gaussian distributed with zero mean and covariance $0.01 \mI$.
We consider a sparse network of Granger causal interactions with only $30 \%$ of the regression coefficients in $\mA^{i}$ selected uniformly being non-zero and the regression matrices~$\mA^{i}$ being collectively joint sparse (same setup as in \cite{Bolstad11GroupLASSOGCI}). All non-zero regression coefficients are set equal to $0.0994$ which guarantees the stability of the simulated VAR process. 

From Table~\ref{tab:var_AUROC}, we observe that all time series models generally achieve a higher AUROC as the number of measurements available increases. For $T = 500$, the component-wise MLP and the proposed eSRU are statistically 
tied when comparing their average AUROCs. 
For $T = 1000$, eSRU significantly outperforms the rest of the time series models and is able to recover the true Granger causal network almost perfectly. %Interestingly, the MLP model is found to outperform the more general LSTM model, possibly due to the latter not regularized enough to model the overly simplistic linear VAR model without overfitting. 
\begin{table}[h]
\begin{minipage}[t]{0.44\textwidth}
	\begin{center}
		\begin{small}
			\caption{Averaged AUROC for $5$ independently generated VAR datasets}
			\label{tab:var_AUROC}
			\begin{tabular}{lccc}
				\multicolumn{1}{c}{\bf MODEL}  &\multicolumn{2}{c}{\bf AVERAGE AUROC}
				\\ \hline \vspace{-0.2cm} \\
				& \bf $T = 500$ & \bf $T = 1000$ & 
				\\ \hline \\
				%MLP 	& 	\bm{$0.939 \pm 0.032$} 	&	$0.925 \pm 0.023$ \\
				%LSTM 	&	$0.787 \pm 0.12$	&	$0.796 \pm 0.088$ \\
				%TCDF    &   $0.765 \pm 0.066$	& 	$0.781 \pm 0.036$ \\
				%SRU		& 	$0.824 \pm 0.06$	& 	$0.911 \pm 0.041$ \\
				%eSRU	& 	$0.927 \pm 0.051$	&	\bm{$0.98 \pm 0.008$} \\
				MLP 	& 	\bm{$0.94 \pm 0.03$} 	&	$0.93 \pm 0.02$ \\
				LSTM 	&	$0.79 \pm 0.12$	&	$0.8 \pm 0.09$ \\
				TCDF    &   $0.77 \pm 0.07$	& 	$0.78 \pm 0.04$ \\
				SRU 		& 	$0.82 \pm 0.06$	& 	$0.91 \pm 0.04$ \\
				eSRU	& 	$0.93 \pm 0.05$	&	\bm{$0.98 \pm 0.01$} \\
			\end{tabular}
		\end{small}
	\end{center}
\end{minipage}
\hspace{0.8cm}
\begin{minipage}[t]{0.49\textwidth}
	\begin{center}
		\begin{small}
			\caption{Average AUROC corresponding to inferred brain connectivity for $5$ human subjects}
			\label{tab:netsim_AUROC}
			\begin{tabular}{lc}
				\bf MODEL  & \bf AVERAGE AUROC
				\\ \hline \vspace{-0.2cm} \\
						& \bf $T = 200$  
				\\ \hline \\
				MLP 	& 	$0.81 \pm 0.04$ \\
				LSTM 	& 	$0.70 \pm 0.03$  \\
				TCDF    & 	$0.75 \pm 0.04$  \\
				SRU 	& 	$0.78 \pm 0.02$  \\
				eSRU 	& 	\bm{$0.84 \pm 0.03$} 	\\
			\end{tabular}
		\end{small}
	\end{center}
\end{minipage}
\vspace{-0.2cm}
\end{table}

\vspace{-0.1cm}
\subsection{In Silico estimation of brain connectivity using BOLD signals}
\vspace{-0.1cm}
In the third set of experiments, we apply the different learning methods to estimate the connections in the human brain from simulated \textit{blood oxygenation level dependent} (BOLD) imaging data. Here, the individual components of $\rvx$ comprise $T$ = $200$ time-ordered samples of the BOLD signals simulated for $n = 15$ different brain regions of interest (ROIs) in a human subject. To conduct the experiments, we use simulated BOLD time series measurements corresponding to the five different human subjects (labelled as $2$ to $6$) in the \textit{Sim-$3$.mat} file shared at \url{https://www.fmrib.ox.ac.uk/datasets/netsim/index.html}. The generation of the \textit{Sim3} dataset is described in \cite{Smith11NetSim}.
The goal here is to detect the directed connectivity between different brain ROIs in the form of pairwise Granger causal relationships between the components of $\rvx$.   

From Table~\ref{tab:netsim_AUROC}, it is evident that eSRU is more robust to overfitting compared to the standard SRU and detects the true Granger causal relationships more reliably. 
Interestingly, a single-layer cMLP model is found to outperform more complex cLSTM and attention gated-CNN (TCDF) models; however we expect the latter models to perform better when more time series measurements are available. 

\vspace{-0.1cm}
\subsection{Dream-3 in silico network inference challenge}
\vspace{-0.1cm}
In the final set of experiments, we evaluate the performance of the different 
time series models in inferring gene regulation networks synthesized for the 
DREAM-$3$ In Silico Network Challenge (\cite{Prill10Dream3, 
Marbach09Dream3DataGen}). Here, the time series $\rvx$ represents the 
\textit{in silico} measurements of the gene expression levels of $n = 100$ 
genes, available for estimating the gene regulatory networks of \textit{E.coli} 
and \textit{yeast}. A total of five gene regulation networks are to be inferred 
(two for E.coli and three for yeast) from the networks' gene expression level 
trajectories recorded while they recover from $46$ different perturbations 
(each trajectory has $21$ time points). %Measurements corresponding to $46$ 
%different perturbations are available for inference. 
All NN/RNN models are implemented with $10$ neurons per layer, except for the componentwise MLP model which has $5$ neurons per layer. 
\begin{table}[h]
\begin{small}
\caption{AUROCs for the inferred gene regulatory networks}
\label{tab:gene_AUROC}
\begin{center}
\begin{tabular}{lccccc}
	\multicolumn{1}{c}{\bf MODEL}  &\multicolumn{5}{c}{\bf AUROC}
	\\ \hline \vspace{-0.2cm} \\
	&  E.coli-$1$ &   E.coli-$2$ &  Yeast-$1$ &  Yeast-$2$ &   Yeast-$3$ 
	\\ \hline \\
	MLP 	& 	$0.644$ &		$0.568$ &		$0.585$ &		$0.506$ &		$0.528$\\
	LSTM 	&	$0.629$	&		$0.609$ &		$0.579$ &		$0.519$ &		$0.555$\\
	TCDF    &   $0.614$	& 		$0.647$ &		$0.581$ &		$0.556$ &		\bm{$0.557$}\\
	SRU		& 	$0.657$	& 		\bm{$0.666$} &	$0.617$ &		\bm{$0.575$} &	$0.55$\\
	eSRU	& 	\bm{$0.66$}	&	$0.629$ &		\bm{$0.627$} &	$0.557$ &		$0.55$\\
\end{tabular}
\end{center}
\end{small}
\vspace{-0.1cm}
\end{table}
From Table~\ref{tab:gene_AUROC}, we can observe that the proposed SRU and eSRU models are generally more accurate compared to the MLP, LSTM, and attention-gated CNN (TCDF) models in inferring the true gene-gene interactions. For four out of the five gene regulatory networks, either SRU or eSRU was the best performing model among the competing ones.

\vspace{-0.1cm}
\section{Conclusion} \label{sec:conclusions}
\vspace{-0.1cm}
In this work, we addressed the problem of inferring pairwise Granger causal relationships between stochastic processes that interact nonlinearly. We showed that the such causality between the processes can be robustly inferred from the regularized internal parameters of the proposed eSRU-based recurrent models trained to predict the time series measurements of the individal processes. 
Future work includes:
%en route to reliable detection of Granger causality between real-world processes from timeseries measurements, e.g., 
%We demonstrated that the underlying Granger causal relationships in a nonlinear dynamical system can be inferred reliably using the model-based inference approach wherein the system's time series measurements in each dimension are fitted with dedicated statistical recurrent units whose regularized parameters encode the true topology of pairwise interactions.    
%Going foward, there is a need to evaluate more principled loss functions besides the mean-squared fit-error loss considered here which can capture the exogenous effects in a more realistic way. Future extensions of this work would consider addressing the following issues:
\begin{enumerate}[i]
	\item Investigating the use of other loss functions besides the mean-square error loss %with a more principled loss function 
	which can capture the exogenous and instantaneous causal effects in a more realistic way.	
	\item Incorporating unobserved confounding variables/processes in recurrent models. 
	\item Inferring Granger causality from multi-rate time series measurements. 
\end{enumerate}

\section*{Acknowledgements}
This work is supported by a Singapore Ministry of Education (MOE) Tier 2 Grant 
(R-263-000-C83-112).

%[SK remark::: explore an architecture which explicitly enforces this ``innovation" viewpoint...interpret scales in $\setA$ as Kalman gain??]

\bibliography{ICLR2020_SRU_for_GCI}
\bibliographystyle{iclr2020_conference}

\ifdefined \SKIPTEMP
\begin{table}[t]
\caption{Sample table title}
\label{sample-table}
\begin{center}
\begin{tabular}{ll}
\multicolumn{1}{c}{\bf PART}  &\multicolumn{1}{c}{\bf DESCRIPTION}
\\ \hline \\
Dendrite         &Input terminal \\
Axon             &Output terminal \\
Soma             &Cell body (contains cell nucleus) \\
\end{tabular}
\end{center}
\end{table}
\fi

\appendix

\section{Review of related work}
Initial efforts in testing for nonlinear Granger causality focused mostly on the nonparameteric approach. \cite{BaekBrock92} proposed a general statistical test to detect nonlinear Granger causality between two variables under the assumption that the linear VAR modeling of their time series measurements results in i.i.d residual errors. The test involves computing correlation integral estimators of the conditional probabilities concerning distances between carefully selected lead-lag subsequences of the input bivariate time series. Successive works by \cite{HiemstraJones94, DiksPanchenko06NGC, Bai10multivarNGC, DiksWolski16NGC} proposed their improved variants of  
the Baek-Brock test. \cite{HiemstraJones94} modified and extended the original Baek-Brock test to allow for weakly dependent residual errors in linear VAR modeling of the time series measurements. %They further showed that 
%their modified test statistic converges asymptotically to the standard 
%Gaussian 
%distribution under the null hypothesis of Granger noncausality. 
\cite{DiksPanchenko06NGC} proposed a rectified version of the Heimstra-Jones' test statistic, making it unbiased while fixing the issue of overrejection of the null hypothesis. %in the Heimstra-Jones (HJ) test. 
%They also showed that their which tends to zero in probability under the null 
%hypothesis of Granger noncausality.   
Later works by \cite{Bai10multivarNGC, DiksWolski16NGC} extended the bi-variate test in \cite{DiksPanchenko06NGC} to the multivariate setting. 
%\textcolor{red}{[Discuss about directed information and transfer entropy based 
%nonparameteric test for Granger causality....]}\textcolor{blue}{[Copula 
%approach for nonlinear Granger causal inference \cite{Hu14CopulaGCI}]}
The biggest common drawback of these nonparameteric tests is the large sample sizes required to robustly 
estimate the conditional probabilities that constitute the test statistic. Furthermore, the prevalent strategy in these methods of testing each one of the variable-pairs individually to detect pairwise Granger causality is unappealing from a computational standpoint, especially when a very large number of variables are involved.
%\cite{BellMalley96NGC}...

In the model driven approach, the Granger causal relationships are inferred 
directly from the parameters of a data generative model fitted to the time series measurements. Compared to the nonparameteric approach, 
the model-based inference approach is considerably more sample 
efficient, however the scope of inferrable causal dependencies is dictated by 
the choice of data generative model. 
%Although linear regression models are  used ubiquitously for testing the existence of Granger causality, they perform poorly in scenarios where the underlying time series data contains nonlinearities. 
Nonlinear kernel based regression models have been found to be reasonably effective in testing of nonlinear Granger causality. Kernel methods rely on linearization of the causal interactions in a kernel-induced 
high dimensional feature space; the linearized interactions are subsequently 
modeled using a linear VAR model in the feature space.  
Based on this idea, \cite{Marrinazzo08KernelGC} proposes a kernel Granger 
causality index to detect pairwise nonlinear Granger causality in the 
multivariate case. %However, the pairwise testing approach is computationally 
%unattractive as the number of pairwise tests grows quadratically with the 
%number of variables. 
In \cite{Sindhwani13KernelGC, Lim14OKVAR}, the nonlinear dependencies in the 
time series measurements are modeled using nonlinear functions expressible as 
sums of vector valued functions in the induced reproducing kernel Hilbert space 
(RKHS) of a matrix-valued kernel.
%also proposes using sums of matrix-valued kernel based RKHS functions for 
%modeling nonlinear causal interactions. 
In \cite{Lim14OKVAR}, additional smoothness and structured sparsity constraints 
are imposed on the kernel parameters to promote consistency of the 
time series fitted nonlinear model.
\cite{Shen16KSVARM} proposes a nonlinear kernel-based structural VAR model to 
capture instantaneous nonlinear interactions. The existing kernel based 
regression models are restrictive as they consider only additive 
linear combinations of the RKHS functions to approximate the nonlinear 
dependencies in the time series. Furthermore, deciding the optimal order of 
kernel based regression models is difficult as it requires prior knowledge of 
the mimimum time delay beyond which the causal influences are negligible. 

By virtue of their universal approximation ability, RNNs offer a 
pragmatic way forward in modeling of complex nonlinear dependencies in the time 
series measurements for the purpose of inferring Granger causality. 
Mutiple recent works by \cite{Wang19RNNGC, Duggento19EchoStateForGCI, Abbasvandi19RNN_GCI_whiteness} 
have investigated the use of different types of RNNs for inferring nonlinear Granger causal relationships. However, they all adopt the same na\"ive strategy whereby each pairwise 
causal relationship is tested individually by estimating its causal connection 
strength. The strength of the causal connection from series $j$ to series $i$ 
is determined by the ratio of mean-squared prediction errors incurred by 
\textit{unrestricted} and \textit{restricted} RNN models towards predicting 
series $i$ using the past measurement sequences of all $n$ component including 
and excluding the $j^{\text{th}}$ component alone, respectively. The pairwise 
testing strategy however does not scale well computationally as the number of 
component series becomes very large. This strategy also fails to exploit the 
typical sparse connectivity of networked interactions between the processes 
which has unlocked significant performance gains in the existing linear methods 
(\cite{Bahadori13, Bolstad11GroupLASSOGCI}).  

In a recent work by \cite{Tank14NeuralGC}, the pairwise Granger causal relationships are 
inferred directly from the weight parameters of component-wise MLP or LSTM 
networks fitted to the time series measurements. 
By enforcing column-sparsity of the input-layer weight matrices in the fitted 
MLP/LSTM models, their proposed approach returns a sparsely connected estimate 
of the underlying Granger causal network.
Due to its feedforward architecture, a traditional MLP network is not 
well-suited for modeling ordered data such as a time series. 
\cite{Tank14NeuralGC} demonstrated that the MLP network can learn short range 
temporal dependencies spanning a few time delays by letting the network's input 
stage process multi-lag time series data over sliding windows. However, 
modeling long-range temporal dependencies using the same approach requires 
a larger sliding window size which entails an inconvenient increase in the 
number of trainable parameters. 
The simulation results in \cite{Tank14NeuralGC} indicate that MLP models are 
generally outperformed by LSTM models in extracting the true topology of 
pairwise Granger causality, especially when the processes interact in 
a highly nonlinear and intricate manner. While purposefully designed for modeling 
short and long term temporal dependencies in a time series, the LSTM (\cite{Hochreiter97LSTM}) is very general and often too much overparameterized and thus prone to overfitting. 
While using overparameterized models for inference is preferable when there is abundant training data 
available to leverage upon, there are several applications where the data available for 
causal inference is extremely scarce. 
It is our opinion that using a simpler RNN model combined with meaningful regularization of the model 
parameters is the best way forward in inferring Granger causal relationships from underdetermined time series measurements. 
%Recently, \cite{Tank14NeuralGC} proposed a general framework for inferring pairwise Granger causality using  feedforward or recurrent neural network based time series prediction models.
Building on the ideas put forth by \cite{Tank14NeuralGC}, this paper investigates the use of Statistical Recurrent Units (SRUs) towards inferring Granger causality.

\section{Proximal gradient descent updates for estimating the regularized weight parameters in the SRU and $e$SRU models}
Noting that the proximal operator corresponding to mixed $\ell_{1}$-$\ell_{2}$ norm (group-norm) is the group-wise soft-thresholding operator, we use the following proximal gradient-descent updates to minimize the $i^{\text{th}}$ SRU's regularized loss in \eqref{eqn_SRU_basic_opt}: 
\begin{align}
\mW_{\mathrm{in}}^{(i),t+1}(:,j) = S_{\lambda_{1}\eta} \lb \mW_{\mathrm{in}}^{(i),t}(:,j) - \eta  \nabla_{\mW_{\mathrm{in}}^{(i)}(:,j)} l_{i}(\Theta^{(i), t}_{\text{SRU}})  \rb, \;\; \forall j \in [n].
\end{align}
Here, $l_{i}(\Theta_{i}^{{t}}) \triangleq \frac{1}{T-1} \sum_{t = 1}^{T-1} \lb \hat{\rx}_{i,t} - \ervx_{i,t+1} \rb^{2}$ is the unregularized SRU loss function, $\eta$ is the gradient-descent stepsize and $S_{\lambda_{1} \eta}$ is the elementwise soft-thresholding operator defined below.
\begin{equation}
S_{\lambda_{1} \eta}(\vw) \triangleq 
\begin{cases}
\vw - \lambda_{1} \eta \frac{\vw}{\| \vw \|_{2}}, & \| \vw \|_{2} > \lambda_{1} \eta \\
0 , &  \| \vw \|_{2} \le \lambda_{1} \eta
\end{cases}, \;\;\;\; \forall \vw \in \Real^{n}. 
\end{equation}
The columns of weight matrix $\mW_{\mathrm{i}}^{(i)}$ in the $i^{\text{th}}$  
eSRU model are also updated in exactly the same fashion as above. 

Likewise, the $j^{\text{th}}$ row of the group-norm regularized weight matrix $\mW_{\mathrm{o}}^{(i)}$ in the eSRU optimization in \eqref{eqn_SRU_opt_mod2} is updated as shown below.
\begin{align}
\mW_{\mathrm{o}}^{(i),t+1}(j,\setG_{j,k}) = S_{\lambda_{2}\eta} \lb \mW_{\mathrm{o}}^{(i),t}(j,\setG_{j,k}) - \eta \nabla_{\mW_{\mathrm{o}}^{(i)}(j,\setG_{j,k})} l_{i}(\Theta^{(i),t}_{\text{eSRU}})  \rb, \;\; \forall j = 1, 2, \ldots d_{\mathrm{o}}.
\end{align} 
The gradient of the unregularized loss function $l_{i}, 1 \le i \le n$ associated with the SRU and eSRU models used in the above updates is evaluated via the backpropagation through time (BPTT) procedure (\cite{Jaeger02TutBPTT}). 

\section{Ablation Study}
\subsection{Choice of encoder for 
$e$SRU feedback's stage-$1$: Fixed or data 
dependent}
As a possible further enhancement of the proposed eSRU time series model, one 
may consider learning the encoding map, $\mD_{r}^{(i)}$, in 
the feedback path, as trainable parameters of the 
$i^{\text{th}}$ eSRU.  
In Table~\ref{tab:ablation_on_D}, we compare the 
Granger causality detection performance of this 
particular eSRU variant and the 
proposed design wherein~$\mD_{r}^{(i)}$ is taken to be 
a random matrix with i.i.d.\ Gaussian entries. 
The experimental setup is kept the same as in  
Section~\ref{sec:sim_results}, and the entries of 
$\mD_{r}^{(i)}$ in the eSRU variant are 
$\ell_{2}$-norm penalized during training.

\begin{table}[h]
\begin{center}
\begin{small}
\caption{Average AUROC for eSRU variants}
\label{tab:ablation_on_D}
\begin{tabular}{lcc}
	\multicolumn{1}{c}{\bf DATASET}  
	&\multicolumn{2}{c}{\bf Average AUROC}
	\\ \hline \vspace{-0.2cm} \\
			& \bf Randomly constructed 
			$\mD_{\mathbf{r}}^{(i)}$   & \bf 
			$\mD_{\mathbf{r}}^{(i)}$ as trainable 
			parameters
	\\ \hline \\
	Lorenz ($T = 250, F= 10$) 	& 	$0.95 \pm 
	0.02$ 		& $0.97 \pm 0.01$ \\
	Lorenz ($T = 500, F= 10$)  	& 	$0.98 \pm 
	0.01$ 		& $0.99 \pm 0.0$ \\
	Lorenz ($T = 250, F= 40$)   & 	$0.99 \pm 
	0.0$  		& $0.98 \pm 0.01$ \\
	Lorenz ($T = 500, F= 40$)	& 	$1.0 \pm 
	0$		 		& $1.0 \pm 0.0$  \\
	VAR ($T = 500$) 			& 	$0.93 \pm 
	0.05$			& $0.91 \pm 0.04$ \\
	VAR ($T = 1000$)    		&   $0.98 \pm 
	0.01$			& $0.98 \pm 0.01$ \\
	NetSim 						&   $0.84 \pm 
	0.03$			& $0.80 \pm 0.02$\\ 
\end{tabular}
\end{small}
\end{center}
\end{table}
We observe that the performance of these two models is statistically tied, 
which indicates that the randomly constructed~$\mD^{(i)}_{\mathrm{r}}$ is able 
to distill the necessary information from the high-dimensional summary 
statistics $\vu_{i,t-1}$ required for generating the feedback  
$\vr_{i,t}$. Based on these results, we recommend using the proposed eSRU 
design with its randomly constructed encoding map $\mD_{\mathrm{r}}^{(i)}$, 
because of its simpler design and reduced training complexity.

\subsection[Impact of group-sparse regularization of output weight 
matrix]{Impact of group-sparse regularization of $\mW_{\mathrm{o}}^{(i)}$}

In order to highlight the importance of 
learning time-localized predictive features in detecting Granger causality, 
we compare the following two time series models:
\begin{enumerate}[i.]
\item proposed eSRU (with group-sparse regularization of 
$\mW_{\mathrm{o}}^{(i)}$ as described in Section~\ref{sec:sru_mod2})
\item eSRU variant with ridge-regularized 
$\mW_{\mathrm{o}}^{(i)}$  
\end{enumerate}
Once again, we use the same experimental settings as mentioned in 
Section~\ref{sec:sim_results}. From Table~\ref{tab:ablation_on_Wo}, we observe 
that barring the \texttt{Lorenz-$96$($T$=$250/500$,$F$=$40$)} datasets, for 
which nearly perfect recovery of the Granger causal network is achieved, the 
average AUROC improves consistently for the other datasets by switching 
from unstructured ridge regularization to the proposed group-sparse 
regularization of the output weight matrix $\mW_{\mathrm{o}}^{(i)}$. 
\begin{table}[h]
\begin{center}
\begin{small}
\caption{Performance of $e$SRU variants with different regularizations of 
$\mW_{\mathrm{o}}^{(i)}$}
\label{tab:ablation_on_Wo}
\begin{tabular}{lcc}
\multicolumn{1}{c}{\bf DATASET}  
&\multicolumn{2}{c}{\bf Average AUROC}
\\ \hline \vspace{-0.2cm} \\
& \bf Proposed group-sparse  & \bf Ridge 
regularization \\
& \bf regularization of $\mW_{\mathrm{o}}^{(i)}$ & 
\bf of $\mW_{\mathrm{o}}^{(i)}$\\
\hline \\
Lorenz ($T = 250, F= 10$) 	& 	$0.95 \pm 0.02$ 		& $0.93 \pm 0.03$ \\
Lorenz ($T = 500, F= 10$)  	& 	$0.98 \pm 0.01$ 		& $0.94 \pm 0.04$ \\
Lorenz ($T = 250, F= 40$)   & 	$0.99 \pm 0.0$  		& $0.99 \pm 0.0$ \\
Lorenz ($T = 500, F= 40$)	& 	$1.0 \pm 0.0$	 		& $1.0 \pm 0.0$  \\
VAR ($T = 500$) 			& 	$0.93 \pm 0.05$			& $0.90 \pm 0.03$ \\
VAR ($T = 1000$)    		&   $0.98 \pm 0.01$			& $0.96 \pm 0.01$ \\
NetSim 						&   $0.84 \pm 0.03$			& $0.83 \pm  0.03$\\ 
\end{tabular}
\end{small}
\end{center}
\end{table}

\section{Implementation details}

\begin{itemize}
	\item \textbf{Activation function for SRU and eSRU models} \\
	While the standard SRU proposed by \cite{Oliva17SRU} uses ReLU neurons, we found in our numerical experiments that using the Exponential Linear Unit (ELU) activation resulted in better performance. The ELU activation function is defined as
	\begin{equation}
	ELU(x) = 
	\begin{cases}
	x & x > 0 \\
	\alpha (e^{x} - 1) & x \le 0 
	\end{cases}
	,\;\; \alpha > 0.     
	\label{defn_elu}
	\end{equation}
	In our simulations, the constant $\alpha$ is set equal to one.
	
	\vspace{0.25cm}		
	\item \textbf{Number of neural layers in SRU model} \\
	To approximate the generative functions $f_{i}$ in \eqref{ts_abstract_generative_model}, we consider the simplest architecture for the SRU networks, whereby the constituent ReLU networks generating the recurrent features, output features and feedback have a single layer feedforward design with equal number of neurons.    
	
	\vspace{0.25cm}		
	\item \textbf{Number of neural layers in Economy-SRU model} \\
	The ReLU networks used for generating the recurrent and output features in the proposed eSRU model have a single-layer feedforward design.
	However, the second stage of eSRU's modified feedback can be either single or multi-layered feedforward network. Provided that $d_{\mathrm{r}}^{\prime} \ll m d_{\mathrm{\phi}}$, a multi-layer implementation of the second stage of eSRU's feedback can still have fewer trainable parameters overall compared to the SRU's single layer feedback network. 
	The simulation results in Section~\ref{sec:sim_results} are obtained using a two-layer ReLU network in the second stage of eSRU's feedback for the DREAM-$3$ experiments, and while using a three-layer design for the Lorenz-$96$, VAR and NetSim experiments.
	
	\vspace{0.25cm}		
	\item \textbf{Self-interactions in Dream-$3$ gene networks} \\
	The in-silico gene networks synthesized for the DREAM-$3$ challenge have no self-connections. Noting that none of the Granger causal inference methods evaluated in our experiments intentionally suppress the self-interactions, the reported AUROC values are computed by ignoring any self-connections in the inferred Granger causal networks.

	%\vspace{0.25cm}		
	%\item \textbf{Smaller and staggered batches for training} \\
	
\end{itemize}

\section{Python codes}
\textbf{cMLP \& cLSTM models} \\
Pytorch implementation of the componentwise MLP and LSTM 
models are taken from \url{https://github.com/icc2115/Neural-GC}. 

\textbf{Temporal Causal Discovery Framework (TCDF) } \\
Pytorch implementation of the attention-gated CNN based Temporal Causal Discovery Framework 
(TCDF) is taken from \url{https://github.com/M-Nauta/TCDF}. 

\textbf{Proposed SRU and Economy-SRU models} \\
Pytorch implementations of the proposed componentwise SRU and eSRU models are shared at 
\url{https://github.com/sakhanna/SRU_for_GCI}.
%\url{https://figshare.com/s/54bae94b7cdad09e20bd}. 

\section{ROC plots}
\begin{figure}[ht]
	\centering
	\begin{subfigure}{.45\textwidth}
		\centering
		\includegraphics[width=\linewidth]{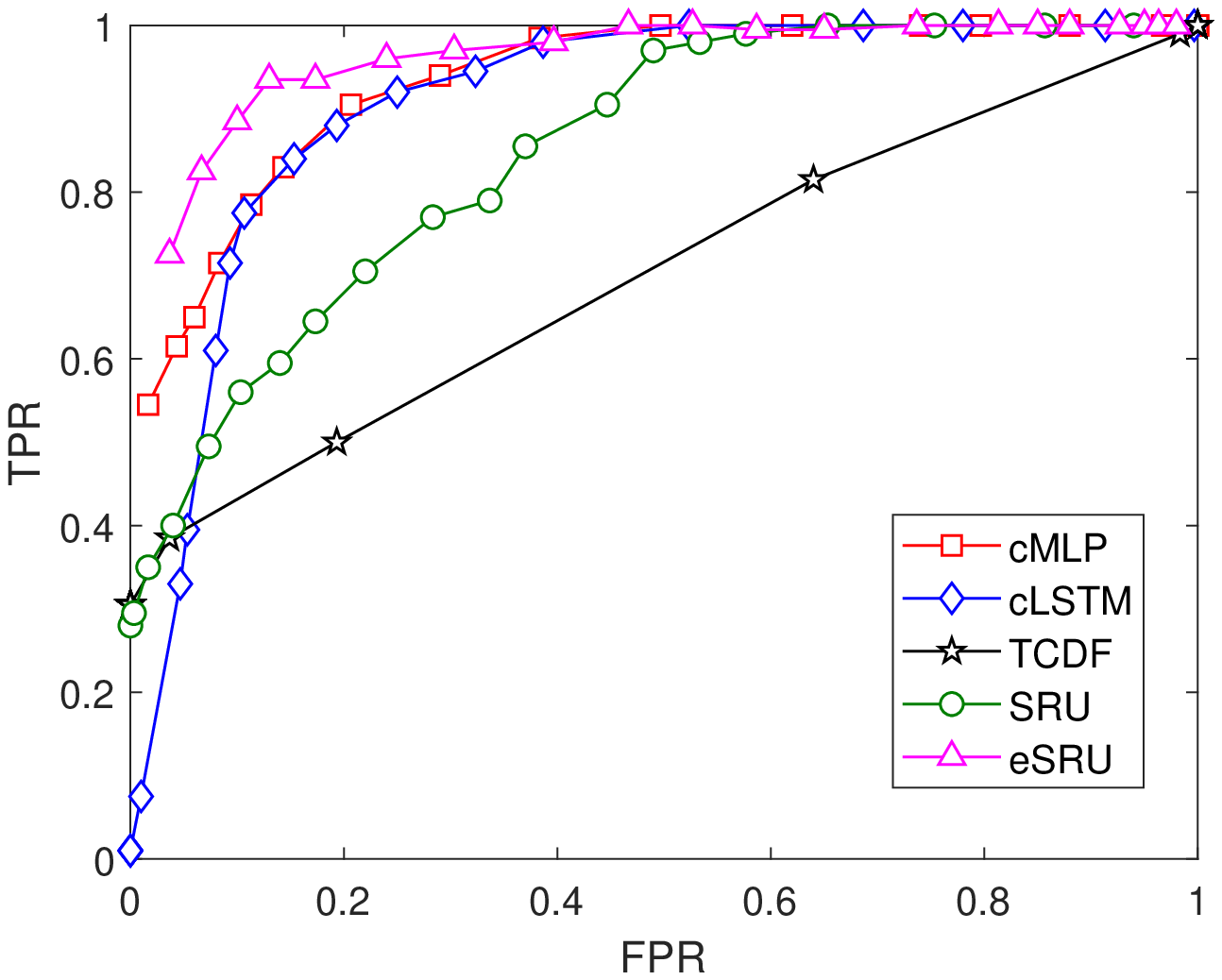}
		\caption{$F$ = $10$, $T$ = $250$ }
		\label{fig:lorenz_F10_T250}
	\end{subfigure}%
	\begin{subfigure}{.45\textwidth}
		\centering
		\includegraphics[width=\linewidth]{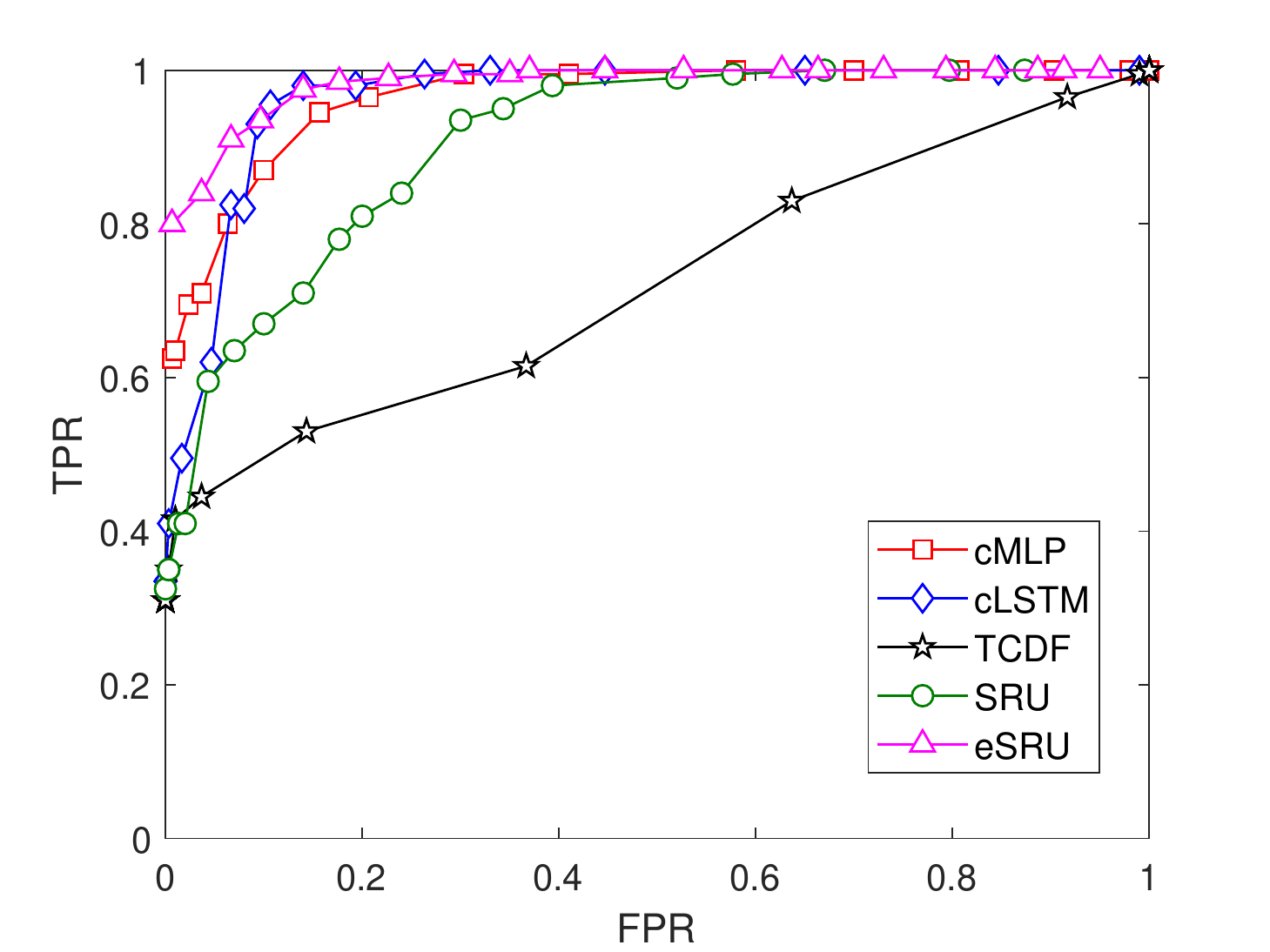}
		\caption{$F$ = $10$, $T$ = $500$}
		\label{fig:lorenz_F10_T500}
	\end{subfigure}
	
	\begin{subfigure}{.45\textwidth}
		\centering
		\includegraphics[width=\linewidth]{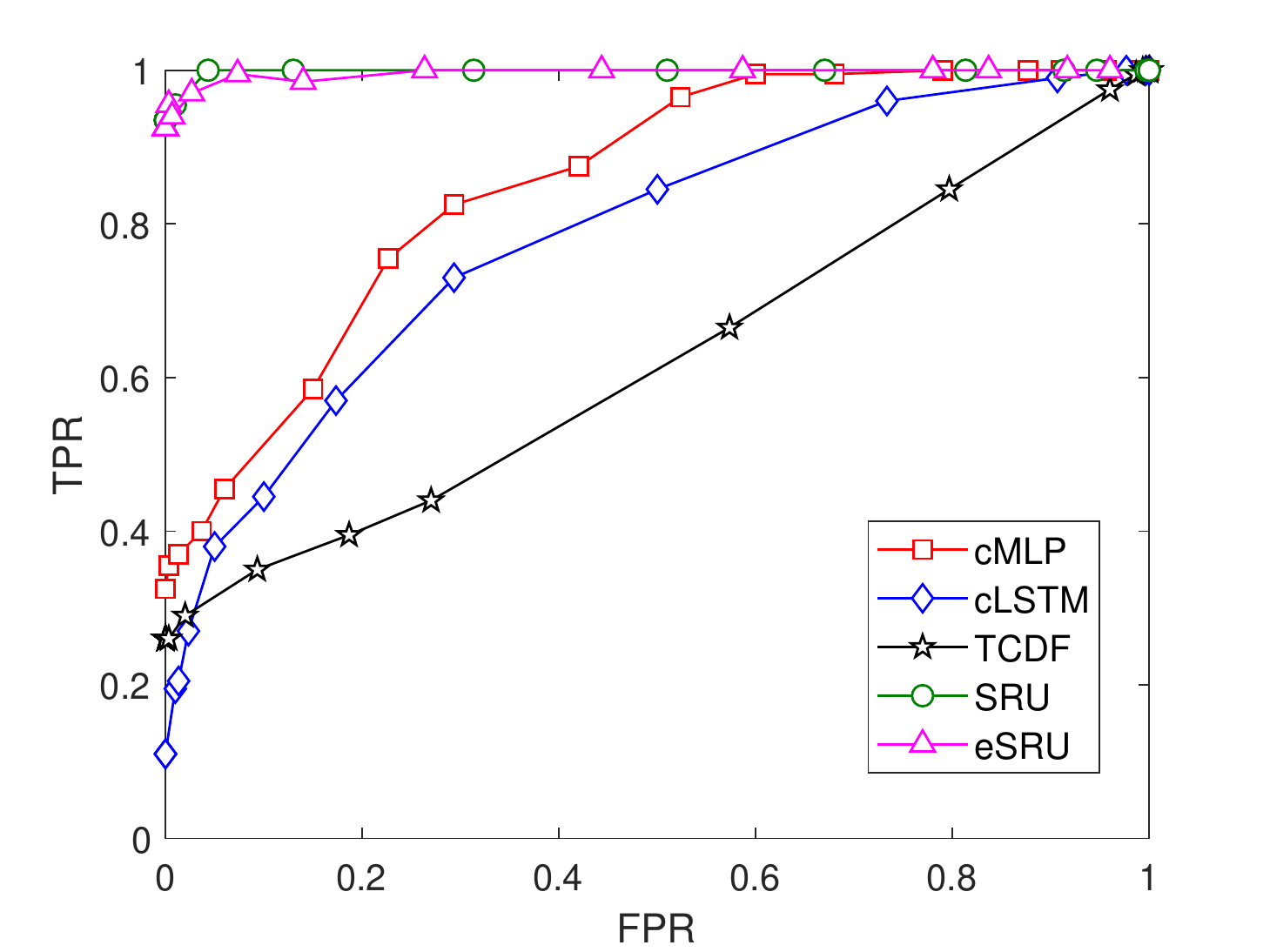}
		\caption{$F$ = $40$, $T$ = $250$}
		\label{fig:lorenz_F40_T250}
	\end{subfigure}%
	\begin{subfigure}{.45\textwidth}
		\centering
		\includegraphics[width=\linewidth]{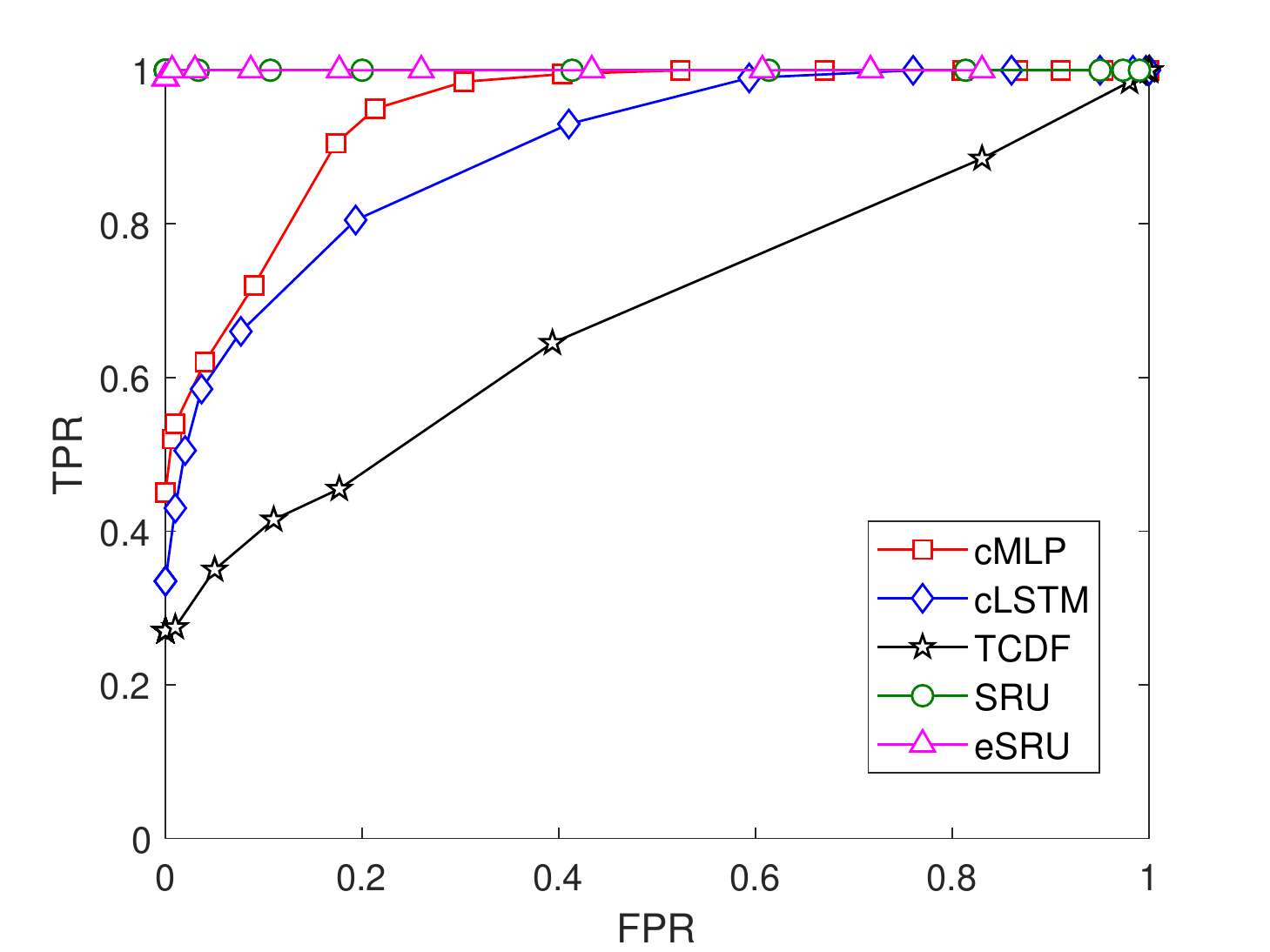}
		\caption{$F$ = $40$, $T$ = $500$}
		\label{fig:lorenz_F40_T500}
	\end{subfigure}
	\caption{Average ROC curves for Lorenz-96 datasets}
	\label{fig:Lorenz_ROC}
\end{figure}

\begin{figure}[h]
	\centering
	\begin{subfigure}{.45\textwidth}
		\centering
		\includegraphics[width=\linewidth]{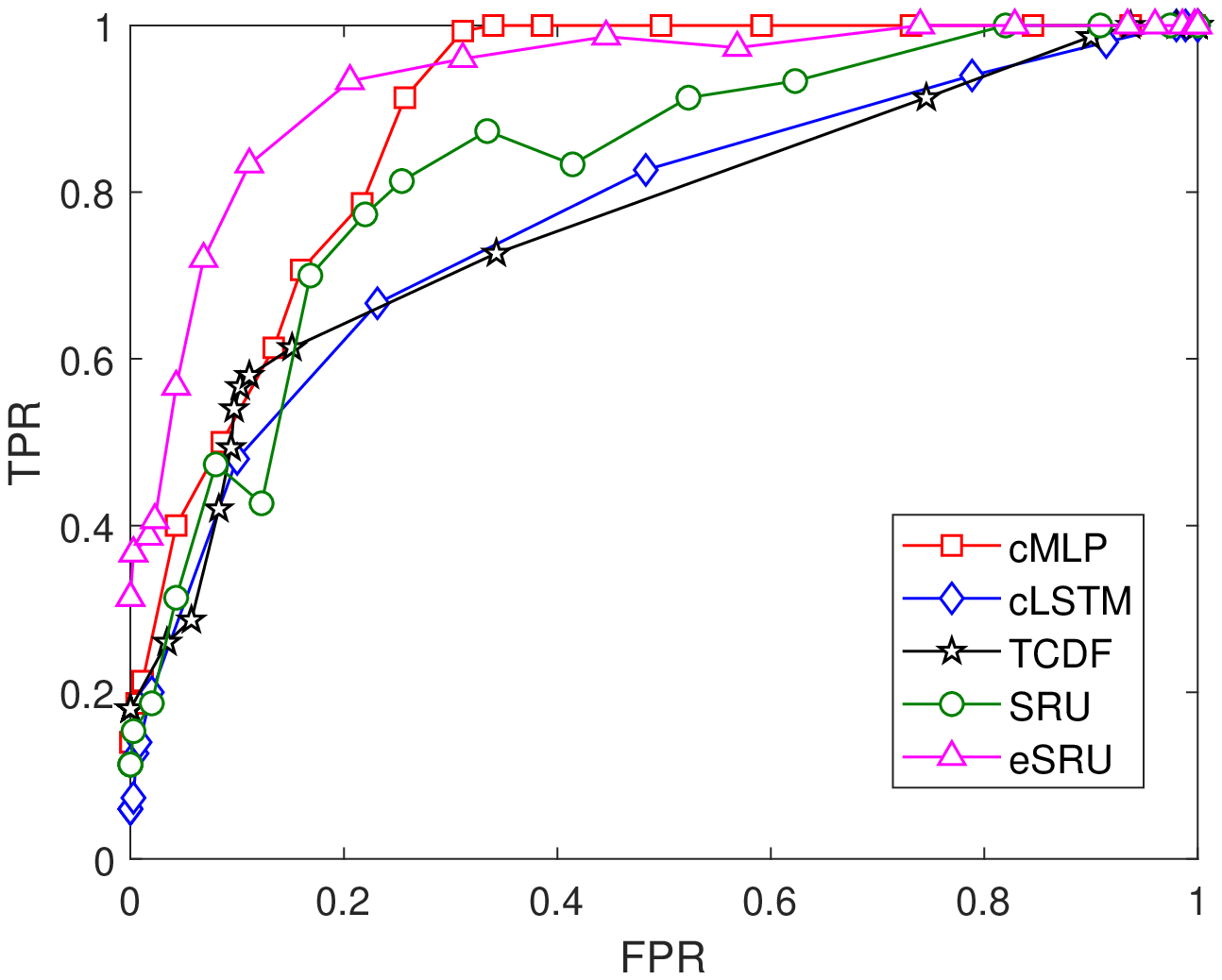}
		\caption{$30 \%$ sparsity, $T$ = $500$ }
		\label{fig:var_T500}
	\end{subfigure}%
	\begin{subfigure}{.45\textwidth}
		\centering
		\includegraphics[width=\linewidth]{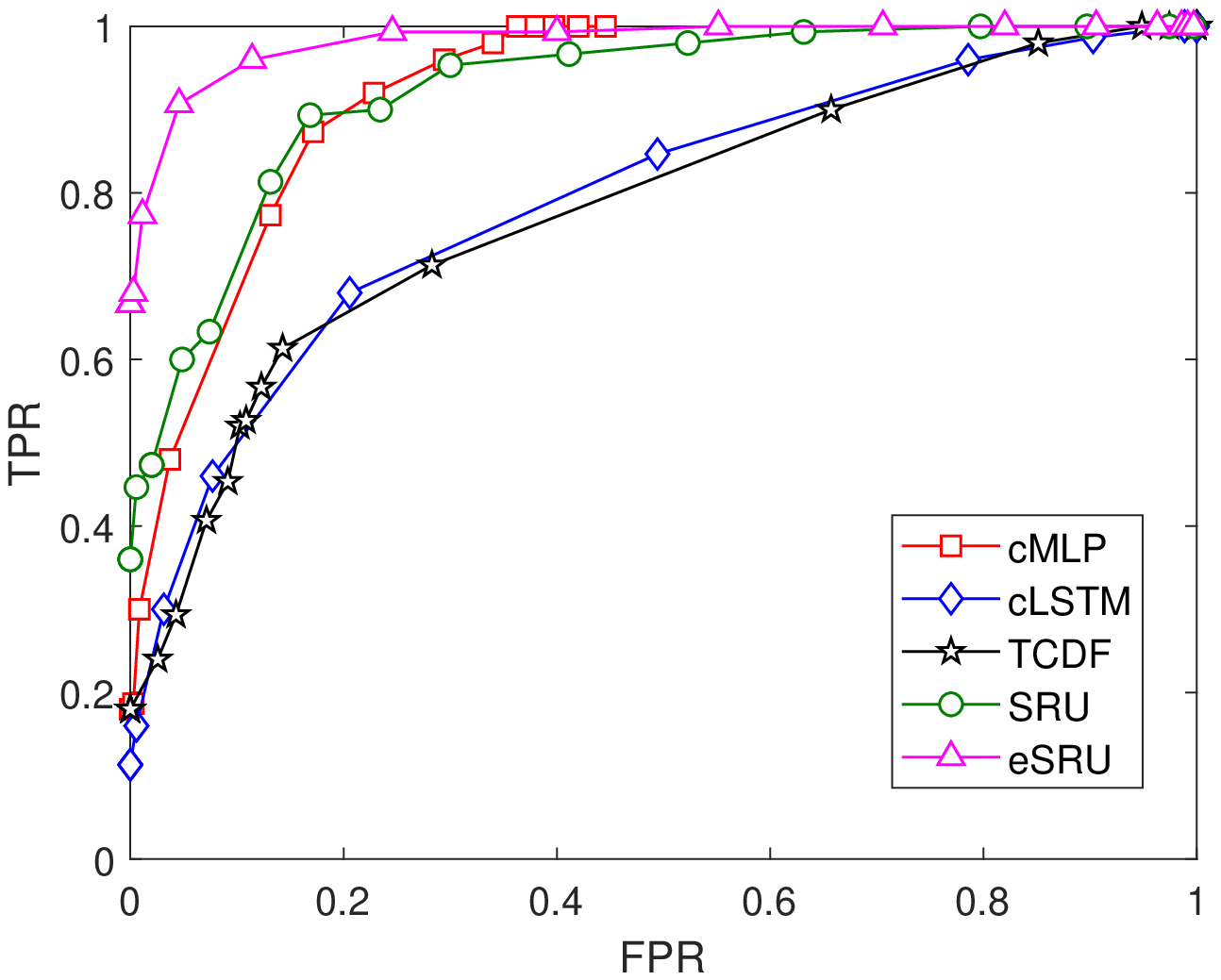}
		\caption{$30 \%$ sparsity, $T$ = $1000$}
		\label{fig:var_T1000}
	\end{subfigure}
	\caption{Average ROC curves for VAR datasets}
	\label{fig:var_roc}
\end{figure}

\begin{figure}[h]
	\centering
	\begin{subfigure}{.45\textwidth}
		\centering
		\includegraphics[width=\linewidth]{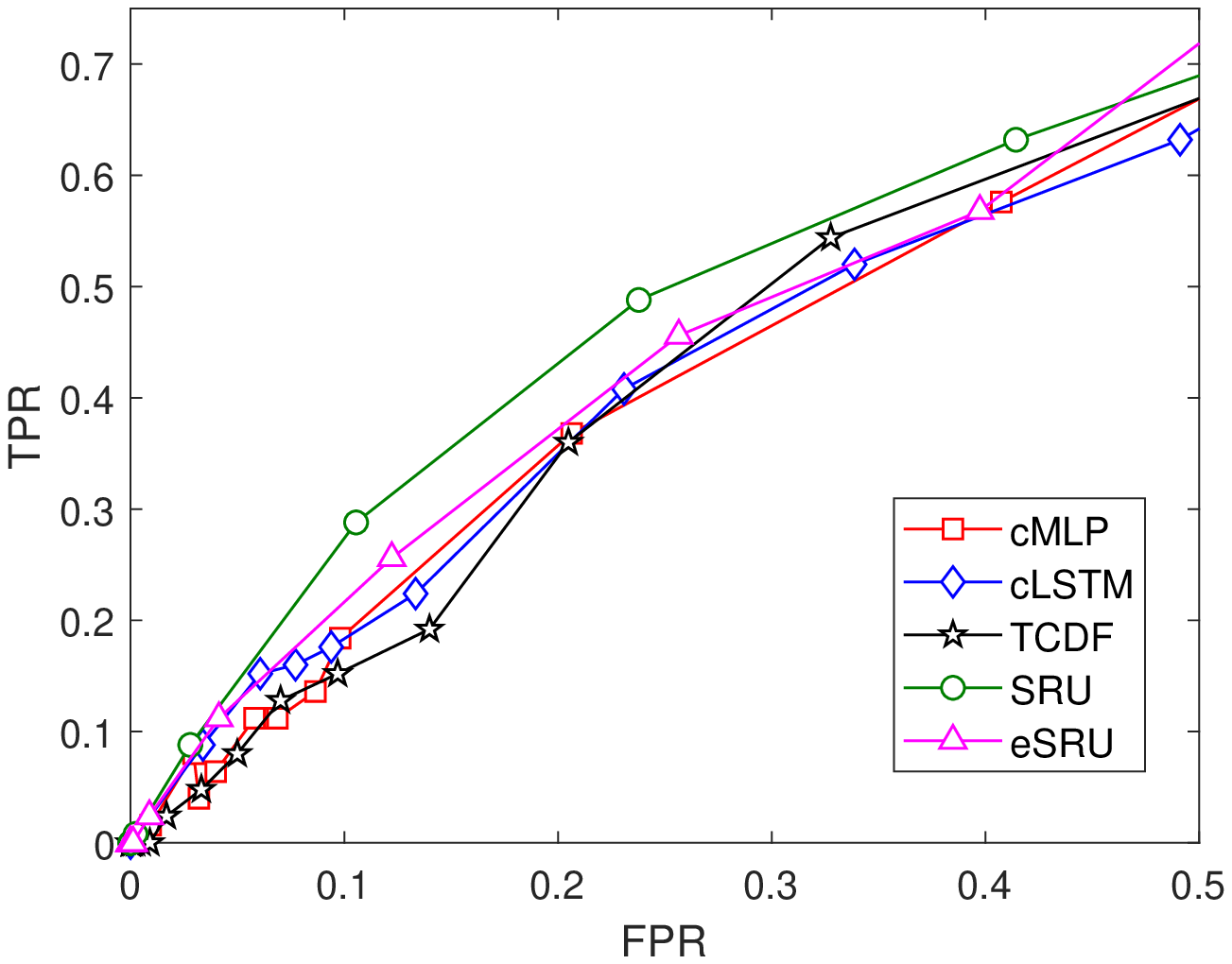}
		\caption{E.coli-$1$}
		\label{fig:ecoli1}
	\end{subfigure}%
	\begin{subfigure}{.45\textwidth}
		\centering
		\includegraphics[width=\linewidth]{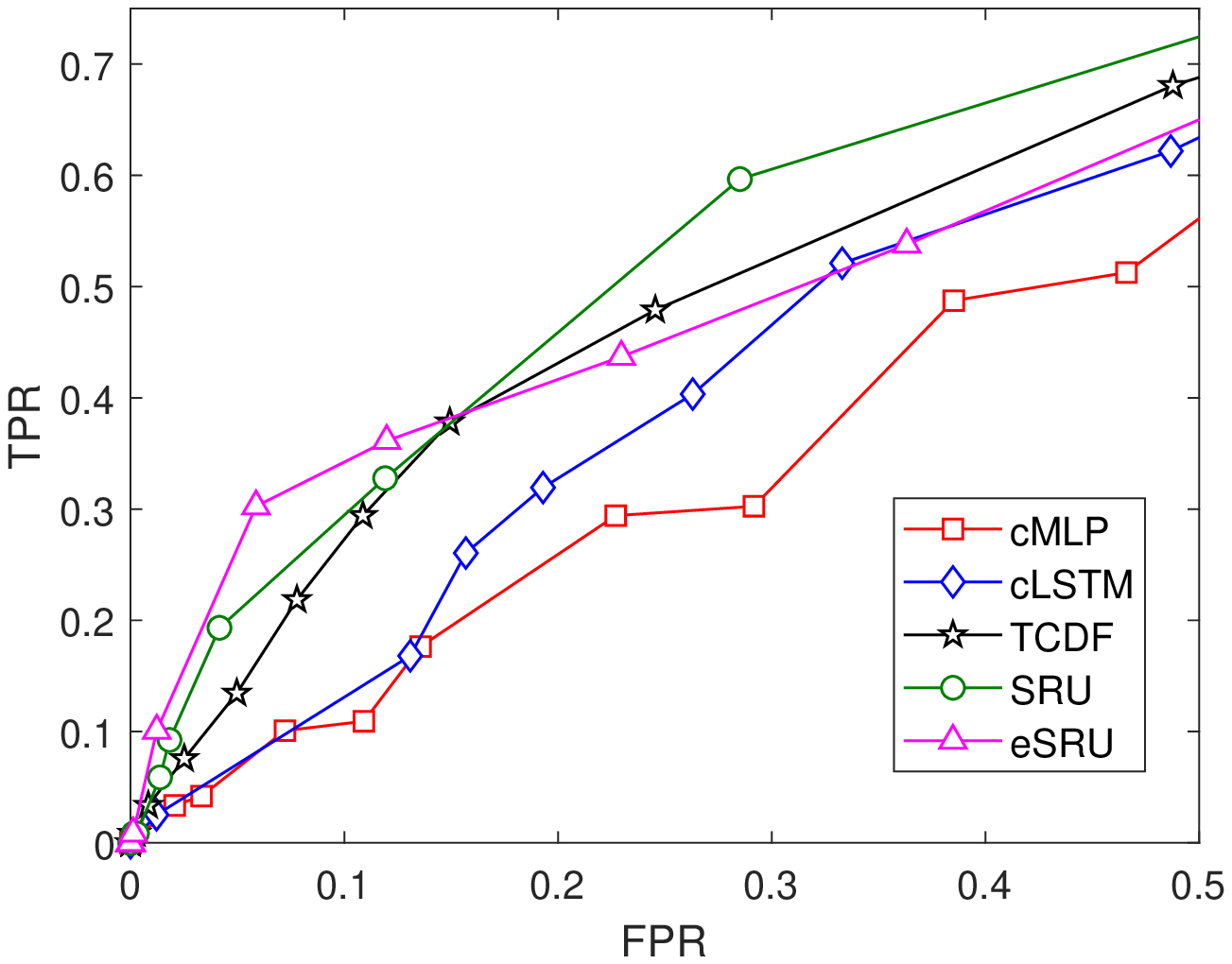}
		\caption{E.coli-$2$}
		\label{fig:ecoli2}
	\end{subfigure}
	
	\begin{subfigure}{.45\textwidth}
		\centering
		\includegraphics[width=\linewidth]{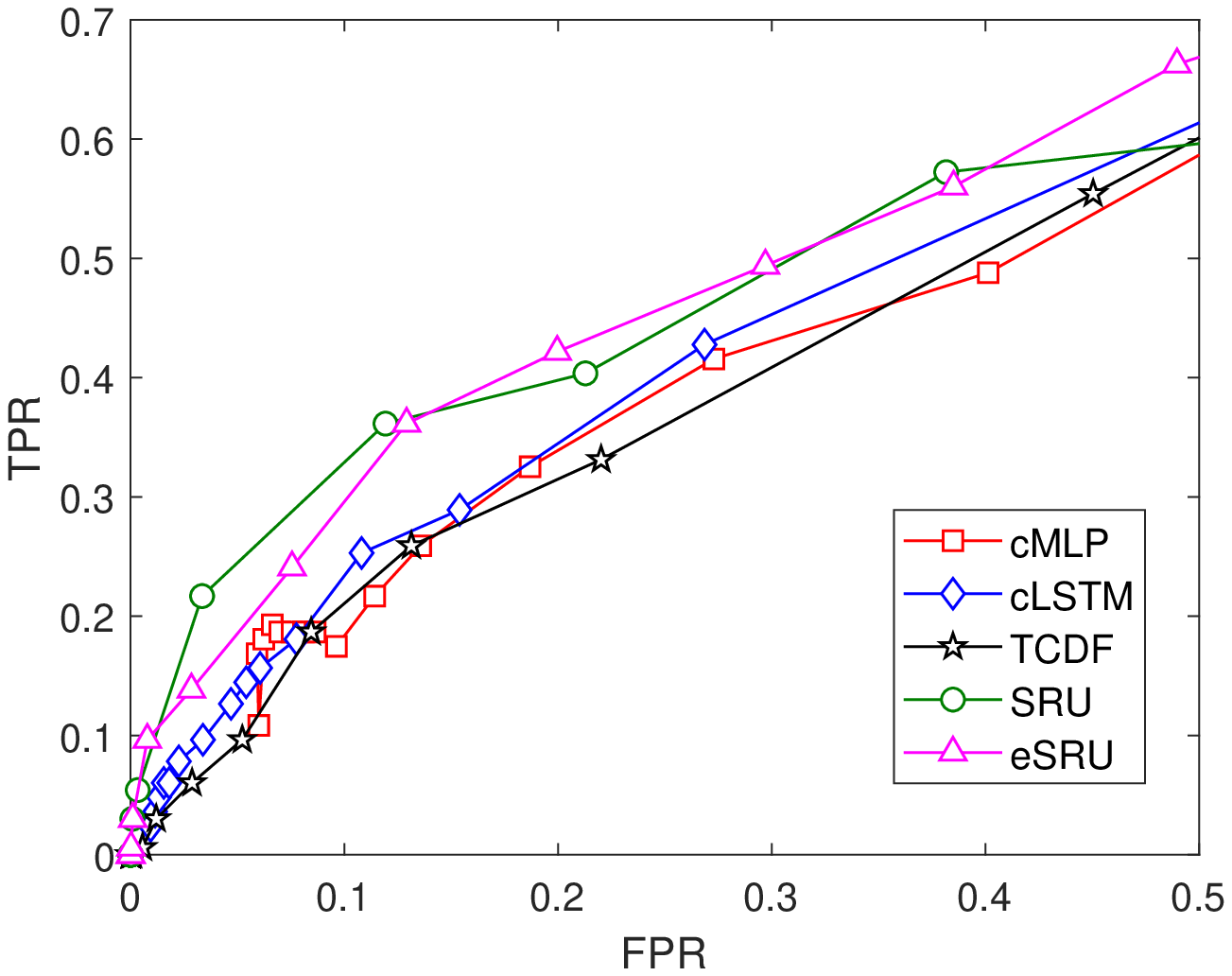}
		\caption{Yeast-$1$}
		\label{fig:yeast1}
	\end{subfigure}%
	\begin{subfigure}{.45\textwidth}
		\centering
		\includegraphics[width=\linewidth]{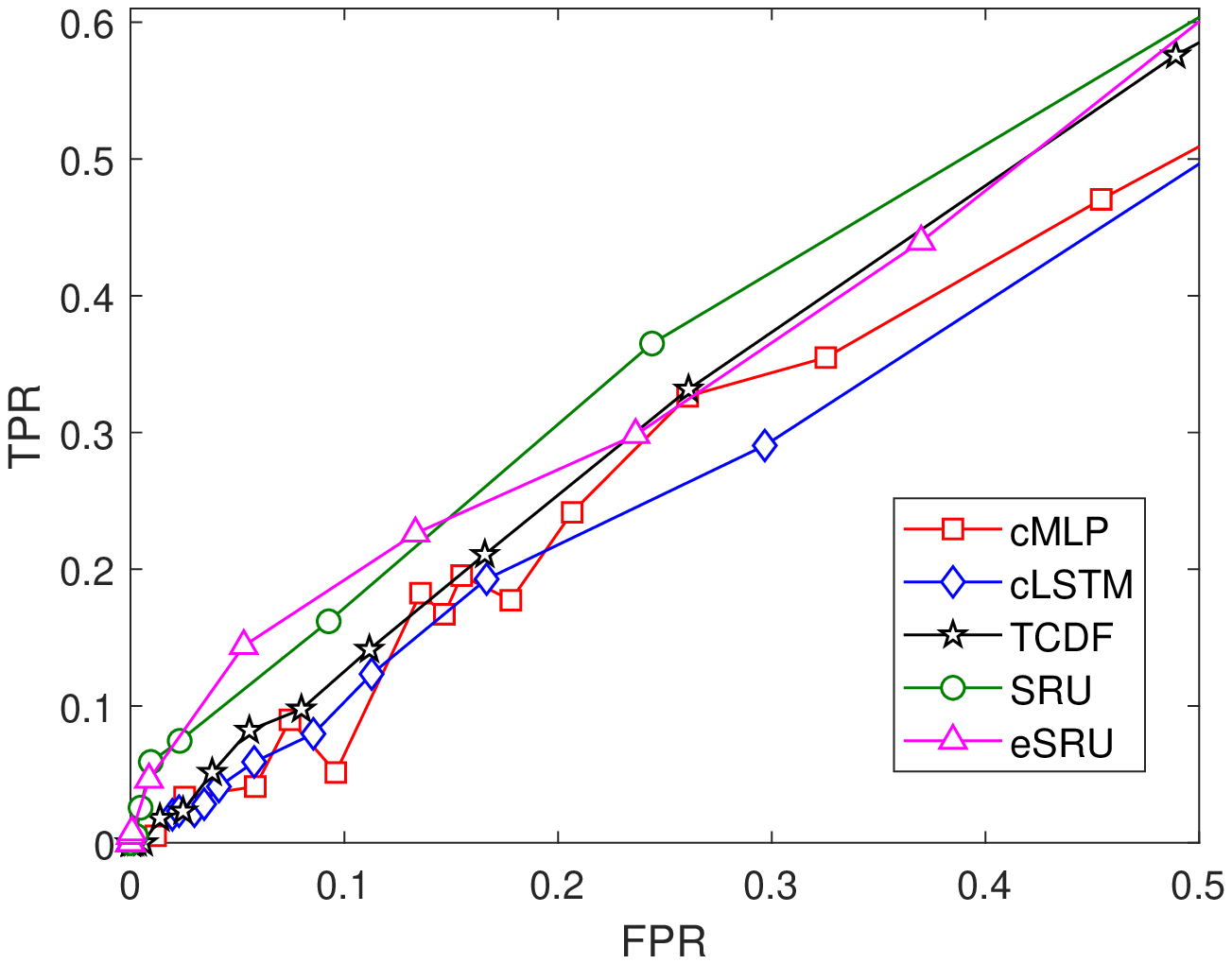}
		\caption{Yeast-$2$}
		\label{fig:yeast2}
	\end{subfigure}
	
	\begin{subfigure}{.45\textwidth}
		\centering
		\includegraphics[width=\linewidth]{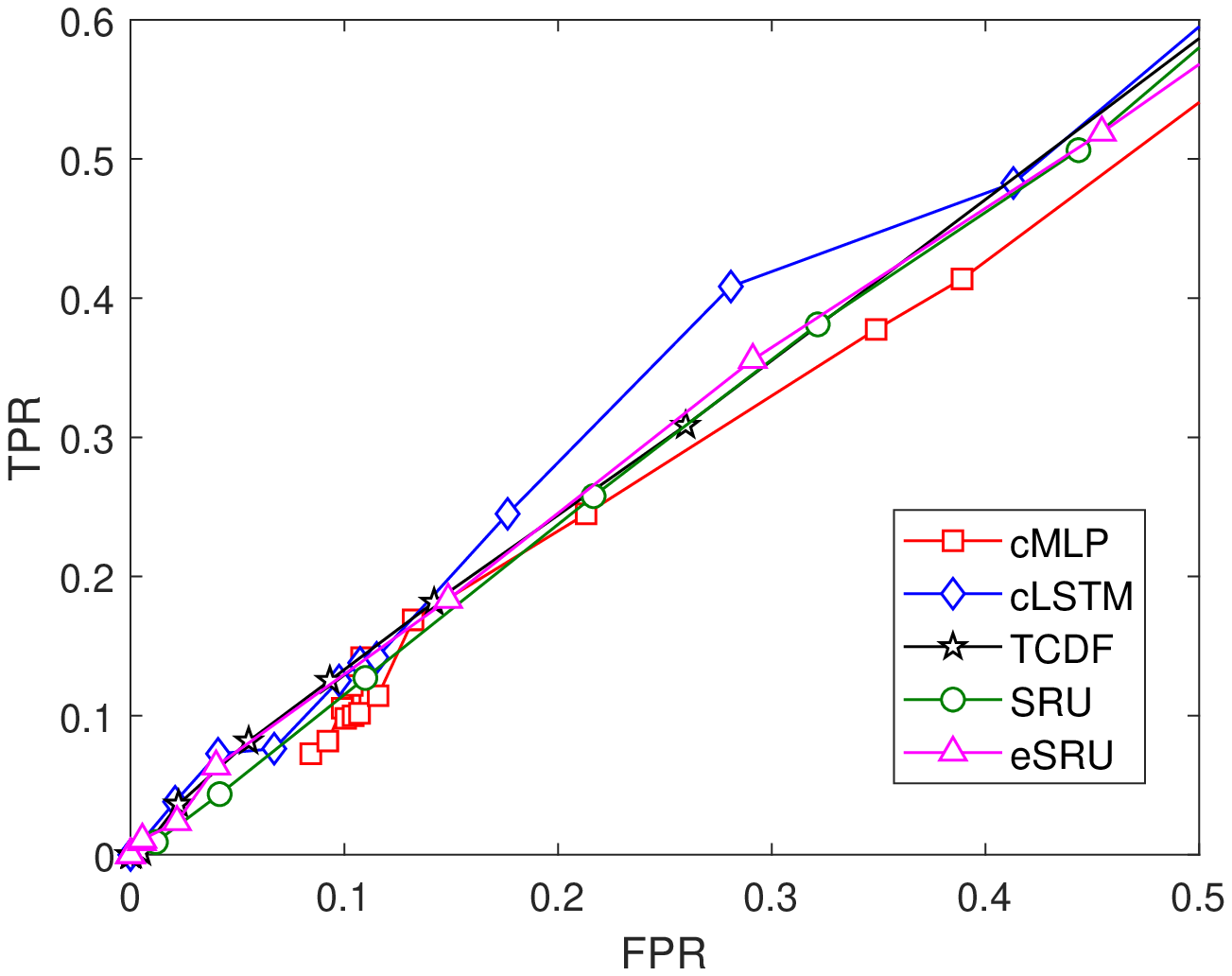}
		\caption{Yeast-$3$}
		\label{fig:yeast3}
	\end{subfigure}
	\caption{ROC curves for Dream-$3$ datasets}
	\label{fig:Dream3_ROC}
\end{figure}

The receiver operating characteristics (ROC) of different Granger causal inference methods 
are compared in Figures~\ref{fig:Lorenz_ROC}-\ref{fig:Dream3_ROC}. %for the simulated Lorenz-$96$ and VAR datasets, respectively. % Each ROC curve is averaged over 5 independently generated datasets. 
Here, an ROC curve represents the trade-off between the true-positive rate (TPR) and the false-positive rate (FPR) achieved by a given method while inferring the underlying pairwise Granger causal relationships. 

\begin{figure}[h]
	\centering
	\includegraphics[width=0.45\linewidth]{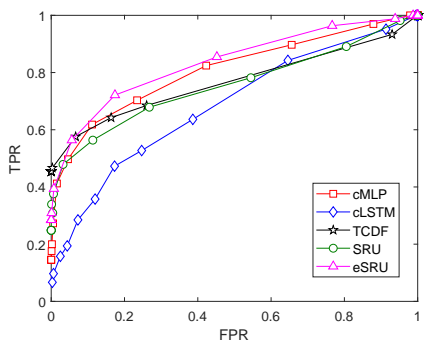}
	\caption{Average ROC curves for the NetSim experiment.}
	\label{fig:var_T500}
\end{figure}

\clearpage

\section{Tuned hyperparameters} \label{app:tuned_hyperparams}
Tables~\ref{tab:mlp_params} to \ref{tab:esru_params} summarize the chosen hyperparameters and configurations of the different NN/RNN models used for generating the results reported in Section~\ref{sec:sim_results}. For the Dream-$3$ experiments, the model configurations used for inferring the \textit{E.coli-$1$} gene regulatory network  \cite{Prill10Dream3} have been provided in the tables.  

\begin{table}[h]
	\begin{scriptsize}
		\centering
		\renewcommand{\arraystretch}{1.2}
		\begin{tabular}{|p{2cm}|p{1.8cm}|c|c|c|c|}
			\hline
			\multirow{ 2}{*}{\textbf{Parameters}} & \multirow{ 2}{*}{\textbf{Tuning range}} & \multicolumn{4}{c|}{\textbf{Dataset}}  
			\\ \cline{3-6}
			& & \bf Lorenz (F = $10/40$) & \bf VAR & \bf Dream-$3$ & \bf NetSim 
			\\ \hline
			$\#$ Neural units per layer & \centering $-$					
			& $10$ 		& $10$ 				& $5$ 	& $10$ 
			\\ \hline 
			Batch size &	\centering $-$				
			& $250/500/1000$ 			& $500/1000$ 				& $21$ 	& 
			$200$ 		
			\\ \hline
			Learning rate & 	Two-fold cross-validation across $[5e\text{-}5, 1e\text{-}1]$					
			& \vtop{\hbox{\strut $0.0005$ ($F$ = $10$),}  \hbox{\strut $0.001$ ($F$ = $40$)} } & 
			$0.1$ 			&  	$0.0005$	& $0.0005$
			\\ \hline
			Ridge regularization bias &  Two-fold cross-validation across $[5e\text{-}5, 10]$ 			
			& \vtop{\hbox{\strut $0.232079$ ($F$ = $10$),}  \hbox{\strut $10.0$ ($F$ = $40$)} } & $0.002$ 			& 	$5.0$	& $0.464159$
			\\ \hline
			Block-sparse regularization bias (input layer) & \hfil	$-$
			& \vtop{\hbox{\strut $[0.1, 10]$ ($F$ = $10$),}  \hbox{\strut $[1, 100]$ ($F$ = $40$)} } & 
			$[0.0001, 0.01]$	& 	$[0.1, 100]$	& $[0.1, 3.162]$
			\\ \hline
			$\#$ Lags & \centering $-$
			& $5$  & $5$ & $2$ & $5$
			\\ \hline
			$\#$ Training epochs & \centering $-$
			& $2000$ & $2000$ & $2000$ & $2000$
			\\ \hline
		\end{tabular}
		\caption{Componentwise MLP model configuration \label{tab:mlp_params} (Refer \cite{TankFox11Nips} for detailed description of the model parameters).}
	\end{scriptsize}
\end{table}

\begin{table}[h]
	\begin{scriptsize}
		\centering
		\renewcommand{\arraystretch}{1.2}
		\begin{tabular}{|p{2cm}|p{1.8cm}|p{1.8cm}|p{1.8cm}|p{1.8cm}|p{1.8cm}|}
			\hline
			\multirow{ 2}{*}{\textbf{Parameters}} & \multirow{ 2}{*}{\textbf{Tuning strategy}} & \multicolumn{4}{c|}{\textbf{Dataset}}   
			\\ \cline{3-6}
			& & \bf Lorenz (F = $10/40$) & \bf VAR & \bf Dream-$3$ & \bf NetSim 
			\\ \hline 
			$\#$ units per layer &	\centering $-$				
			& $10$ 			& $10$ 				& $10$ 	& $10$ 
			\\ \hline 
			Batch size &	\centering $-$				
			& $250/500/1000$ 		& $500/1000$ 	& $21$ 	& $200$ 			
			\\ \hline
			Learning rate &	Two-fold cross-validation across $[5e\text{-}5, 1e\text{-}1]$							
			& \vtop{\hbox{\strut $0.0005$ ($F$ = $10$),}  \hbox{\strut $0.001$ ($F$ = $40$)} }	& $0.1$ 			&  $0.0005$		& $0.001$
			\\ \hline
			Ridge regularization bias 	& Two-fold cross-validation across $[5e\text{-}5, 10]$ 				
			& \vtop{\hbox{\strut $0.021544$ ($F$ = $10$),}  \hbox{\strut $5.0$ ($F$ = $40$)} } & $0.0005$ 			& 	$5.0$	& $0.010772$
			\\ \hline
			Group-sparse regularization bias &	\centering $-$
			& $[1,56.234]$ 	& $[0.003162, 0.01]$	& $[0.1, 17.52]$		& $[0.1, 3.162]$
			\\ \hline
			Truncation & \centering $-$
			& No		 		& No				 & No 	& No 
			\\ \hline
			$\#$ Training epochs & \centering $-$
			& $2000$ &  $2000$ & $2000$ & $4000$
			\\ \hline
		\end{tabular}
		\caption{Componentwise LSTM model configuration \label{tab:lstm_params} (Refer \cite{TankFox11Nips}) for detailed description of the model parameters).}
	\end{scriptsize}
\end{table}

\begin{table}[h]
	\begin{scriptsize}
		\centering
		\renewcommand{\arraystretch}{1.2}
		\begin{tabular}{|p{2cm}|p{2.2cm}|c|c|c|c|c|}
			\hline
			\multirow{ 2}{*}{\textbf{Parameters}} & \multirow{ 2}{*}{\textbf{Tuning strategy}}  & \multicolumn{5}{c|}{\textbf{Dataset}}   
			\\ \cline{3-7}
			& & \bf Lorenz (F $= 10$) & \bf Lorenz (F $= 40$) & \bf VAR & \bf 
			Dream-$3$ & \bf NetSim
			\\ \hline 
			Kernel size & \centering	$\{2,4\}$				
			& $4$ 			& $2$		& $2$ 				& $2$ 	& $4$ 
			\\ \hline 
			Batch size & \centering	$-$				
			& $250/500/1000$ 			& $250/500/1000$		& 
			$500/1000$ 				& $21$ 	& $200$ 			\\ \hline
			Layers & \centering $\{2,3,4\}$							
			& $2$ 		& $2$	& $3$ 			&  $3$		& $2$
			\\ \hline
			Learning rate 	& \centering $\{ 10^{-1}, 10^{-2}, 10^{-3} \}$		
			& $0.01$ 	& $0.01$ 		& $0.001$ 			& 	$0.01$	& $0.001$
			\\ \hline
			Dilation &	\centering $\{1,2,4\}$
			& $1$ 	& $2$	& $1$	& 	$4$	& $2$
			\\ \hline
			Significance & \centering $-$
			& $0.8$			& $0.8$ 		& $8$				 & $0.8$ 	& $0.8$ 
			\\ \hline
			$\#$ Training epochs & \centering $\{1000,2000,4000\}$
			& $2000$ & $2000$ & $2000$ & $4000$ & $2000$
			\\ \hline
		\end{tabular}
		\caption{TCDF's Attention-gated CNN model parameters \label{tab:tcdf_params} (Refer \cite{Nauta19TCDF} for detailed description of the model parameters).}
	\end{scriptsize}
\end{table}

\begin{table}[h]
\begin{scriptsize}
	\centering
	\renewcommand{\arraystretch}{1.2}
	\begin{tabular}{|p{2cm}|p{1.8cm}|p{1.8cm}|p{1.8cm}|p{1.8cm}|p{1.8cm}|}
		\hline
		\multirow{ 2}{*}{\textbf{Parameters}} & \multirow{ 2}{*}{\textbf{Tuning strategy}} 
		& \multicolumn{4}{c|}{ \centering \textbf{Dataset}}   
		\\ \cline{3-6}
		& & \bf Lorenz (F = $10/40$)  & \bf VAR & \bf Dream-$3$ & \bf NetSim
		\\ \hline 
		$\#$ units per layer &	\centering $-$				
		& $10$ 			& $10$ 				& $10$ 	& $10$ 
		\\ \hline
		$\setA$ 	& \centering $-$
		& $\{0.0, 0.01, 0.1, $ $ 0.99\}$	& $\{0.0, 0.01, 0.1, $ $ 0.99\}$		& $\{0.0, 0.01, 0.1, $ $ 0.5, 0.99\}$ 	& $\{0.0, 0.01, 0.1, $ $ 0.99\}$
		\\ \hline
		Learning rate 	&	Two-fold cross-validation across $[5e\text{-}4, 1e\text{-}1]$ 						
		& $0.005$ ($F$ = $10$), $0.01$ ($F$ = $40$) &  $0.04$ 			&  $0.005$	& $0.001$
		\\ \hline
		Ridge regularization bias 	&		Two-fold cross-validation across $[0.01, 10]$ 	
		& \vtop{\hbox{\strut $0.021544$ ($F$ = $10$),}  \hbox{\strut $0.464159$ ($F$ = $40$)} }		& $0.021544$ 			& 	$0.2$	& $0.464159$
		\\ \hline
		Group-sparse regularization bias $\lambda_{1}$ (input layer)	& \centering $-$
		& \vtop{\hbox{\strut $[0.1, 1]$ ($F$ = $10$),} \hbox{\strut $[0.0631,1]$ ($F$ = $40$)} } & $[0.001, 1]$	& 	$[0.01, 1.0]$	& $[0.1, 3.162]$
		\\ \hline
		Batch size & \centering $-$
		& $125$ & $125$ & $21$ & $5$
		\\ \hline
		$\#$ Training epochs & \centering $-$
		& $2000$ & $2000$ & $1000$ & $2000$
		\\ \hline
	\end{tabular}
	\caption{Componentwise SRU model configuration \label{tab:sru_params}}
\end{scriptsize}
\end{table}

\begin{table}[h]
\begin{scriptsize}
	\centering
	\renewcommand{\arraystretch}{1.2}
	\begin{tabular}{|p{2cm}|p{1.8cm}|p{1.8cm}|p{1.8cm}|p{1.8cm}|p{1.8cm}|}
		\hline
		\multirow{ 2}{*}{\textbf{Parameters}} & \multirow{ 2}{*}{\textbf{Tuning strategy}} & \multicolumn{4}{c|}{\centering \textbf{Dataset}}   
		\\ \cline{3-6}
		& & \bf Lorenz (F = $10/40$) & \bf VAR & \bf Dream-$3$ & \bf NetSim
		\\ \hline 
		$\#$ units per layer 	&	\centering $-$			
		& $10$ 			& $10$ 				& $10$ 	& $10$ 
		\\ \hline			
		$\setA$ 	& \centering $-$
		& $\{0.0, 0.01, 0.1, $ $ 0.99\}$	& $\{0.0, 0.01, 0.1, $ $ 0.99\}$		& $\{0.05, 0.1, 0.2, $ $ 0.99\}$ 	& $\{0.0, 0.01, 0.1, $ $ 0.99\}$
		\\ \hline
		Learning rate 	&	Two-fold cross-validation across $[5e\text{-}4, 1e\text{-}1]$ 							
		& $0.01$ 		& $0.01$ 			&  	$0.001$	& $0.001$
		\\ \hline
		Ridge regularization parameter 		&	Two-fold cross-validation across $[0.01, 10]$ 		
		& \vtop{\hbox{\strut $0.001$ ($F$ = $10$),} \hbox{\strut $0.043088$ ($F$ = $40$)}} & $0.021544$  & $0.1$		& $0.232$
		\\ \hline
		Group-sparse regularization bias $\lambda_{1}$ (input layer) &	\centering $-$
		& $[0.03162, 0.1]$ 	& $[0.03162, 0.3162]$	& $[0.1, 3.162]$		& $[0.1, 3.162]$
		\\ \hline
		Group-sparse regularization bias $\lambda_{2}$ (output feature layer) 	& Two-fold cross-validation across $[0.01,10]$
		& \vtop{\hbox{\strut $0.232079$ ($F$ = $10$),} \hbox{\strut $0.928318$ ($F$ = $40$)}} 	& $0.464159$	& $1.0$ 	& $0.005$ 
		\\ \hline
		$\#$ layers in the second stage of feedback network 	& \centering $-$
		& $2$	 		& $2$				 & $1$ 	& $2$ 
		\\ \hline
		Batch size & \centering $-$
		& $125$ & $125$ & $21$ & $5$		
		\\ \hline
		$\#$ Training epochs & \centering $-$
		& $2000$ & $2000$ & $2000$ & $2000$
		\\ \hline
	\end{tabular}
	\caption{Economy-SRU model configuration \label{tab:esru_params}}
\end{scriptsize}
\end{table}

\ifdefined \SKIPTEMP
\section{\textcolor{red}{TBD}}
\begin{itemize}
	\item cite Guo18LSTMExoModel
\end{itemize}
\fi

\ifdefined \SKIPTEMP
\subsection{SRU vs other RNN architectures}
It is noteworthy that the Granger causal inference framework proposed by \cite{Tank14NeuralGC} is quite general and easily adapted to work with different RNN architectures. 
Here we highlight some of the key architectural aspects of SRU that makes it a compelling model for  inferring nonlinear Granger causality. 
Compared to LSTM and GRU, SRU has a simpler architecture which can capture both long and short term relations in a time series without using the nonlinear sigmoid gating functions. The memory of any specific past segment of the time series is emulated by taking weighted sums of the multi-timescale summary statistics, each summary statistic having a different level of sensitivity to older portions of the time series.     

...Noise rejection ability....

\begin{itemize}
	\item Compared to SRU, the features are more disentangled and therefore more interpretable..
	\item 1. maximal lag selection is automatic/data driven
	\item IIR filter bank view -- highly interpretable 
	\item avoids use of highly nonlinear activation functions --> conducive to theoretical analysis
	\item Framework extends naturally to matrix and tensor valued hidden states   
	\item SRU generative model is more suited for modelling continuous time signals..
\end{itemize}

It is often the case that the occurrence of an interesting pattern in a time series can be attributed to one or more highly time-localized events that occurred in the past. Such types of causal explainations are naturally modeled by an SRU network.

Gated recurrent units \cite{Chung2014GRU}, LSTM \cite{Hochreiter97LSTM}, echo state RNN 
-- Are bidirectional RNNs ruled out for causal inference?....SRU generalizes GRU cell state update ??...
\fi

\end{document}